\documentclass[a4paper,fleqn]{cas-dc}

\usepackage[authoryear]{natbib}
\usepackage{lineno}
\usepackage{xcolor}
\usepackage{bbm}
\usepackage{subcaption}  
\usepackage{graphicx}
\usepackage[export]{adjustbox}
\usepackage{colortbl}
\usepackage{arydshln} 
\usepackage{amssymb}
\usepackage{float}
\usepackage{placeins}
\usepackage{microtype}

\def\tsc#1{\csdef{#1}{\textsc{\lowercase{#1}}\xspace}}
\tsc{WGM}
\tsc{QE}

\definecolor{myred}{RGB}{200,50,50}
\definecolor{mygreen}{RGB}{34,139,34}

\begin{document}
\let\WriteBookmarks\relax
\def\floatpagepagefraction{1}
\def\textpagefraction{.001}

\shorttitle{Incentivizing Botanical Reasoning in MLLMs with Reinforcement Learning for Precision Weed Grounding}    

\shortauthors{Z. Yang et~al.}  

\title [mode = title]{WeedExpert-R1: Incentivizing Botanical Reasoning in MLLMs with Reinforcement Learning for Precision Weed Grounding}



%

\author[1,2]{Zonglin Yang}






\affiliation[1]{organization={Department of Biological Systems Engineering, University of Nebraska-Lincoln},
            city={Lincoln},
            state={NE},
            country={USA}}

\author[1,2]{Wei-Zhen Liang}[orcid=0000-0001-9454-5682]


\cormark[1]

\ead{wei-zhen.liang@unl.edu}

\author[2,3]{Nevin Lawrence}
\author[1,2]{Xin Qiao}
\author[4]{Benjamin Riggan}
\author[1,2]{Chi-En Chiang}
\author[1,2]{Fuchen Li}




\affiliation[2]{organization={Panhandle Research and Extension Center, University of Nebraska-Lincoln},
            city={Scottsbluff},
            state={NE},
            country={USA}}

\affiliation[3]{organization={Department of Agronomy and Horticulture, University of Nebraska-Lincoln},
            city={Lincoln},
            state={NE},
            country={USA}}

\affiliation[4]{organization={Department of Electrical and Computer Engineering, University of Nebraska-Lincoln},
            city={Lincoln},
            state={NE},
            country={USA}}

\cortext[1]{Corresponding author}




\begin{abstract}
Precision weed control requires both species-level identification and per-instance localization, but conventional object detectors are constrained by a closed-vocabulary paradigm that hinders cross-region deployment and lack the reasoning capability to justify their predictions in complex agricultural scenes. Recently, multimodal large language models (MLLMs) have demonstrated strong visual perception and reasoning in visual grounding, offering a promising alternative. However, their insufficient domain-specific botanical knowledge often leads to hallucinations during fine-grained weed identification. In this study, \textbf{WeedExpert-R1} is introduced, a multimodal reasoning model that, following the R1-style training paradigm, incentivizes visually grounded botanical reasoning through verifiable rewards. Specifically, a domain-specific Chain-of-Thought (CoT) synthesis pipeline is proposed. It pairs a human-curated botanical trait dictionary, covering leaf shape, margin, petiole, and stem morphology, with an Auditor–Synthesizer LLM workflow to generate high-quality reasoning data for supervised fine-tuning as a cold start. This pipeline bridges the gap between textual botanical knowledge and the visual perception of MLLMs. Group Relative Policy Optimization (GRPO) is then applied with verifiable rewards (format, accuracy, count, and length penalty) to further enhance the model's capability to perceive and localize diverse weeds. WeedExpert-R1-4B achieves 75.82\%, 89.30\%, and 87.81\% on Precision@(F$_1$=1, IoU$\ge$0.5), Precision@0.5, and Recall@0.5 across 37 weed species on a benchmark suite of six weed datasets (3SeasonWeedDet10, CottonWeedDet12, CottonWeedDet3, Weed-crop, Weed25, and PREEC), and outperforms both frontier proprietary models (GPT-5.4, Gemini-3.1-Pro) and larger open-source baselines (Qwen3-VL-30B-Instruct, Gemma-4-31B-it). Moreover, qualitative results on unseen species further illustrate the open-vocabulary capability of WeedExpert-R1, suggesting potential for deployment across diverse regions and crops without retraining.
\end{abstract}


\newcommand{\wz}[1]{\textcolor[rgb]{0,0,1}{#1}}
\newcommand{\zl}[1]{\textcolor{ForestGreen}{#1}}

\begin{keywords}
 Precision agriculture\sep Weed grounding\sep Multimodal large language models\sep Reinforcement learning\sep
\end{keywords}

\maketitle

\section{Introduction}

Weeds compete with crops for essential resources, such as water, nutrients, and light, posing a significant threat to crop yields~\citep{soltani2016potential, lawrence2021herbicide}. Herbicide application has long served as the primary method of weed management, but the rising prevalence of herbicide-resistant weed populations is rapidly reducing the effectiveness of conventional chemical control~\citep{miranda2024crop, akuoko2026evaluating}. These challenges create an urgent need for control technologies that reduce chemical inputs while maintaining effective management and that can target the specific resistant species present in a given field.

The herbicide-resistant weeds that drive yield loss differ markedly by crop and by region. In sugar beet production in the western Nebraska panhandle, kochia (\textit{Bassia scoparia}), Palmer amaranth (\textit{Amaranthus palmeri}), and volunteer corn (\textit{Zea mays}) are the dominant resistant species~\citep{lawrence2021herbicide}. In dry edible bean production in the Bighorn Basin of Wyoming, the priority list shifts to Venice mallow (\textit{Hibiscus trionum}), hairy nightshade (\textit{Solanum sarrachoides}), and Palmer amaranth. Even within the same crop, the priority list changes across states and growing regions. Therefore, effective field deployment requires not only species-level identification but also the flexibility to reconfigure which species to target on a per-crop and per-region basis. Once the target species is correctly identified, accurately localizing each weed instance further enables site-specific applications, such as precision spraying and laser weeding, reducing herbicide use and harm to the crop and the surrounding environment.

\begin{figure}[t]
  \centering
    \includegraphics[scale=0.42]{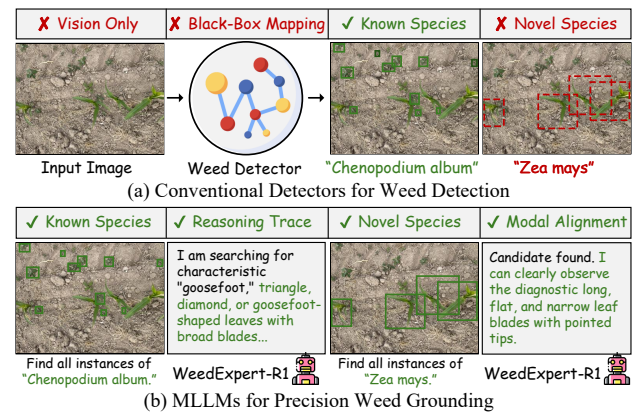}
    \caption{Comparison of conventional weed detection and MLLM-based weed grounding. 
    (a) A conventional detector maps visual features to predictions from a predefined set of categories, provides no explicit reasoning, and is not designed to identify species outside its training categories. (b) WeedExpert-R1, our fine-tuned MLLM, receives the target species as a natural-language query and uses botanical reasoning to identify and localize the requested species. This framework provides a basis for open-vocabulary grounding of species beyond a fixed detector class list.}\label{grounding-detection}
    \vspace{-1em}
\end{figure}

Computer vision provides an automated approach for identifying and localizing every weed instance. Conventional detectors, such as YOLO~\citep{khanam2024yolov11}, DINO~\citep{zhang2022dino}, and
  RT-DETR~\citep{zhao2024detrs}, typically operate in a closed-vocabulary paradigm (Fig.~\ref{grounding-detection}), in which the target species are defined during training time. This paradigm leads to two practical challenges. First, when a detector is deployed in a new region or crop, the priority species list changes, and any species absent from the original training set cannot be detected~\citep{yang2026weedcam,wang2026resource}. Second, even when the species set remains unchanged, domain shift caused by variation in soil, illumination, camera viewpoint, weed growth stage, or canopy density might reduce accuracy because the model has learned only the visual patterns of its training distribution. Addressing either limitation requires collecting and annotating new field data and retraining, which can be costly and time-consuming within a seasonal production cycle. In addition, conventional detectors generally output class labels and bounding boxes without explicitly describing the visual or botanical evidence supporting their predictions, making uncertain results more difficult for growers and agronomists to assess. These limitations motivate a more flexible framework that can accept target species at inference time and provide interpretable evidence for their localization.
   
Multimodal large language models (MLLMs)~\citep{grattafiori2024llama, bai2025qwen3} have recently demonstrated strong image understanding capabilities, which may support agricultural applications that require fine-grained perception and factually grounded responses. For instance, \textsc{AgMMU}~\citep{gauba2026agmmu} evaluates the agricultural capabilities of MLLMs through both multiple-choice questions and open-ended questions. The authors also developed a comprehensive corpus for domain-specific fine-tuning that covers insect pests, diseases, and symptoms. Because expert-annotated agricultural data remain limited, other studies have adapted general-purpose MLLMs using instruction-tuning datasets assembled from existing agricultural image datasets: AgroGPT~\citep{awais2025agrogpt}, for instance, was trained on AgroInstruct, an instruction-tuning corpus assembled from six vision-only datasets, with the expert-tuning data generated using LLaVA\citep{liu2023visual}. More recently, reinforcement learning (RL) has shown promise for improving the reasoning performance of MLLMs~\citep{jaech2024openai, guo2025deepseek}. Agri-R1~\citep{zhang2026agri} extended this paradigm to agriculture by incorporating domain-specific terminology into its reward signal. Their training set is built on CDDMBench~\citep{liu2024multimodal}, a large-scale agricultural Visual Question Answering (VQA) benchmark across 16 crop species and 60 disease categories. Together, these studies demonstrate the value of domain-specific data and training strategies for adapting MLLMs to agricultural applications.

Despite these advances, most agricultural MLLM studies evaluate image-level understanding and fail to localize the objects they describe. Precision weed management, by contrast, requires spatial prediction because interventions such as targeted spraying and robotic removal must operate on individual plants rather than the entire image of the weeds. Visual grounding addresses this need by locating image regions that correspond to a natural-language query~\citep{he2023grec, liu2024grounding}. Unlike conventional closed-vocabulary detectors, a grounding model can receive the target species as an inference-time prompt rather than as a predefined class label (Fig.~\ref{grounding-detection}). Large-scale pretraining equips MLLMs with broad visual and textual knowledge beyond a fixed task-specific training set, providing a foundation for open-vocabulary recognition and grounding. In principle, this allows the same model to be queried for different priority weeds across crops and regions. For example, it could identify kochia, Palmer amaranth, and volunteer corn in a Nebraska sugar beet field, and then Venice mallow, hairy nightshade, and Palmer amaranth in a Wyoming dry bean field, without modifying a predefined set of categories or retraining the model. Additionally, large-scale pretraining also exposes MLLMs to diverse visual conditions, which may improve their ability to generalize across domains, and their step-by-step reasoning capabilities can provide interpretable explanations for their predictions rather than only class labels and bounding boxes. These properties open a path toward open-vocabulary weed grounding: a single, interpretable model deployable across regions and crops.

Realizing this potential, however, requires MLLMs to reliably associate species names with the diagnostic botanical traits used for weed identification. Recent works have explored reinforcement learning with verifiable rewards (RLVR)~\citep{liu2025visionreasoner, liu2025visual} and agentic RL~\citep{zheng2025deepeyes, hong2025deepeyesv2} to improve reasoning and localization in visual-grounding tasks. Despite these advances, general-purpose MLLMs still struggle with reliable weed grounding. In our experiments, even frontier models such as GPT-5.4 ~\citep{openai2026gpt54} and Gemini-3.1-Pro~\citep{deepmind2026gemini31} showed limited performance in species-level grounding (Section~\ref{main-results}). These results suggest that their broad capabilities alone are insufficient to meet the demands of precision weed identification, meaning that although MLLMs are capable of describing many weed species in text, they typically fail to connect species names with diagnostic visual traits that \textit{Weed Scientists} use for identification, such as leaf shape, margin, petiole, and stem morphology. To address this gap, this study presents \textbf{WeedExpert-R1}, a multimodal reasoning model that follows the R1-style training paradigm~\citep{guo2025deepseek}. Botanical reasoning is instilled through supervised fine-tuning (SFT), and species-level weed identification and localization are improved through reinforcement learning with verifiable rewards (RLVR).


The main contributions are fourfold: 1) a domain-specific chain-of-thought (CoT) synthesis pipeline was developed to generate botanically grounded reasoning traces. The pipeline pairs a human-curated trait dictionary with an Auditor–Synthesizer LLM workflow and structures the reasoning process into four stages: trait recall, candidate scanning, contradiction resolution, and instance aggregation. 2) A two-stage post-training procedure was established. Supervised fine-tuning (SFT) makes use of the synthesized CoT data to provide a foundation for structured botanical reasoning, followed by Group Relative Policy Optimization (GRPO)~\citep{shao2024deepseekmath}, which improves species-level grounding through four verifiable rewards. 3) WeedExpert-R1-4B was evaluated across 37 weed species using a benchmark suite comprising five publicly available datasets and one dataset collected at the University of Nebraska–Lincoln Panhandle Research and Extension Center (PREEC). The model achieved 75.82\%, 89.30\%, and 87.81\% on Precision@(F$_1$=1, IoU$\ge$0.5), Precision@0.5, and Recall@0.5, respectively, and outperformed the evaluated proprietary models, including GPT-5.4 and Gemini-3.1-Pro, as well as larger open-source baselines, including Qwen3-VL-30B-Instruct and Gemma-4-31B-it. 4) Qualitative results for selected species excluded from the post-training data provide initial evidence of zero-shot open-vocabulary grounding and demonstrate the model’s potential for adaptation across different crops and regions without retraining on a predefined class set.


\section{Materials and Methods}
\subsection{Reinforcement Learning for Large Language Models}

In a Markov Decision Process defined by a tuple $\mathcal{M} = (\mathcal{S}, \mathcal{A}, \mathcal{P}, r, \gamma)$, the agent takes an action $a_{t} \in \mathcal{A}$ given a state $s_{t} \in \mathcal{S}$ at each time step $t$, receiving a corresponding reward $r(s_{t}, a_{t})$. The state transition function $\mathcal{P}(s_{t+1} | s_{t}, a_{t}): \mathcal{S} \times \mathcal{A} \times \mathcal{S} \rightarrow [0, 1]$ defines the probability of transitioning to state $s_{t+1}$ given the current state $s_{t}$ and action $a_{t}$. The standard reinforcement learning (RL) objective is to find the optimal policy $\pi_{\theta}$ by maximizing the expected cumulative reward:
\begin{equation}
	\mathcal{J}(\theta) = \mathbb{E}_{\tau\sim \pi_{\theta}}\left[ \sum_{t=0}^{T}\gamma^{t}r(s_{t}, a_{t}) \right]\ ,
\end{equation}
where $\tau = (s_{0}, a_{0}, r_{0}, ..., s_{T}, a_{T}, r_{T})$ denotes the trajectory sampled from the policy $\pi_{\theta}$.

In the context of language model generation, a trajectory corresponds to a complete response $o$ to a given prompt $q$. Accordingly, the RL objective simplifies to:
\begin{equation}
\mathcal{J}(\theta) = \mathbb{E}_{q\sim \mathcal{D},o\sim \pi_{\theta}(\cdot | q)}\left[ r(q,o) \right], \ 
\end{equation}
where the prompt $q$, serving as the initial state, is sampled from the data distribution $\mathcal{D}$, and the policy $\pi_{\theta}$ is the language model being optimized.

\textbf{Proximal Policy Optimization (PPO)~\citep{schulman2017proximal}.} PPO introduces a clipped surrogate objective to stabilize policy updates:
\begin{equation}
\label{ppo}
\begin{split}
    \mathcal{J}_{\textrm{PPO}}(\theta) &= \mathbb{E}_{q\sim \mathcal{D},o\sim \pi_{\theta_{old}}(\cdot | q)} 
     \\
    &\frac{1}{\vert o \vert} \sum^{\vert o \vert}_{t=1} \min \Biggl[ \frac{\pi_{\theta}(o_{t} | q,o_{<t})}{\pi_{\theta_{old}}(o_{t} | q,o_{<t})} \hat{A}_{t}, \\
    &\text{clip}\left(\frac{\pi_{\theta}(o_{t} | q,o_{<t})}{\pi_{\theta_{old}}(o_{t} | q,o_{<t})}, 1 - \epsilon, 1 + \epsilon\right)\hat{A}_{t} \Biggr]\ ,
\end{split}
\end{equation}
where $\frac{\pi_{\theta}(o_{t} | q,o_{<t})}{\pi_{\theta_{old}}(o_{t} | q,o_{<t})}$ denotes the importance-sampling ratio, $\epsilon$ is the clipping threshold, $\hat{A}_{t}$ is an estimator of the advantage at time step $t$, and $\pi_{\theta}$ denotes the language-model policy being optimized.

To estimate $\hat{A}_{t}$ with reduced variance, PPO commonly adopts an actor–critic framework with a value model $V_{\phi}$ parameterized by $\phi$. The advantage then can be estimated using generalized advantage estimation (GAE)~\citep{schulman2015high}.

\textbf{Group Relative Policy Optimization (GRPO)~\citep{shao2024deepseekmath}.} GRPO eliminates the need for a separate value model $V_{\phi}$. Instead, it samples a group of responses $\{ o_{i} \}^{G}_{i=1}$ for each prompt and computes corresponding rewards $\{r_{i} \}^{G}_{i=1}$. The advantage for each response $o_{i}$ is then estimated as:
\begin{equation}
	\hat{A}_{i,t} = \frac{r_{i} - \text{mean}\left(\{r_{i} \}^{G}_{i=1}\right)}{\text{std}\left(\{r_{i} \}^{G}_{i=1}\right)}\ .
\end{equation}
GRPO also introduces a KL-divergence penalty into the PPO objective (Eq.~\ref{ppo}):
\begin{equation}
\label{grpo}
\begin{split}
	\mathcal{J}&_{\textrm{GRPO}}(\theta) = \mathbb{E}_{q\sim \mathcal{D},\{o_{i}\}_{i=1}^{G}\sim \pi_{\theta_{old}}(\cdot | q)} \\
	 &\frac{1}{G}\sum_{i=1}^{G}\frac{1}{\vert o_{i} \vert}\sum^{\vert o_{i} \vert}_{t=1} \Biggl\{
	 \min \biggl[ \frac{\pi_{\theta}(o_{i,t} | q,o_{<t})}{\pi_{\theta_{old}}(o_{i,t} | q,o_{<t})}\hat{A}_{i,t},\\
	 &\text{clip}\left(\frac{\pi_{\theta}(o_{i,t} | q,o_{<t})}{\pi_{\theta_{old}}(o_{i,t} | q,o_{<t})}, 1 - \epsilon, 1 + \epsilon\right)\hat{A}_{i,t} \biggr] \\
	  &- \beta\mathbb{D}_{KL}\left[ \pi_{\theta} \Vert \pi_{ref} \right] \Biggl\}\ ,
\end{split}
\end{equation}
where $\pi_{ref}$ is the SFT model, and $\mathbb{D}_{KL}$ is an unbiased estimator of the KL divergence, which can be computed by:
\begin{equation}
	\mathbb{D}_{\textrm{KL}}\left[ \pi_{\theta} \Vert \pi_{ref} \right] = \frac{\pi_{\theta}(o_{i,t} | q,o_{<t})}{\pi_{\theta_{old}}(o_{i,t} | q,o_{<t})} - \log \frac{\pi_{\theta}(o_{i,t} | q,o_{<t})}{\pi_{\theta_{old}}(o_{i,t} | q,o_{<t})} - 1\ .
\end{equation}

\textbf{Reinforcement Learning with Verifiable Rewards (RLVR).} Although GRPO eliminates the need for a value model $V_{\phi}$, a reward model $R$ is still requires reward signals to guide policy optimization. For tasks with objectively verifiable answers, such as code generation, mathematical reasoning, object detection, and visual grounding, rewards can be computed directly from task outcomes rather than learned from human preference data. This gives rise to a reward-model-free RL paradigm that employs verifiable reward functions to evaluate the model response directly. Combining GRPO with verifiable reward functions therefore simplifies the post-training of LLMs and MLLMs.

\textbf{Cold Start.} Although GRPO with verifiable rewards removes the separate value model used in PPO, direct RL optimization can be unstable when the model lacks a structured reasoning pattern. Therefore, \cite{guo2025deepseek} proposed supervised fine-tuning (SFT) using a small set of chain-of-thought (CoT) examples before RL training. High-quality CoT supervision provides a stronger initialization for subsequent RL optimization.

\begin{figure*}[t]
  \centering
    \includegraphics[scale=0.41]{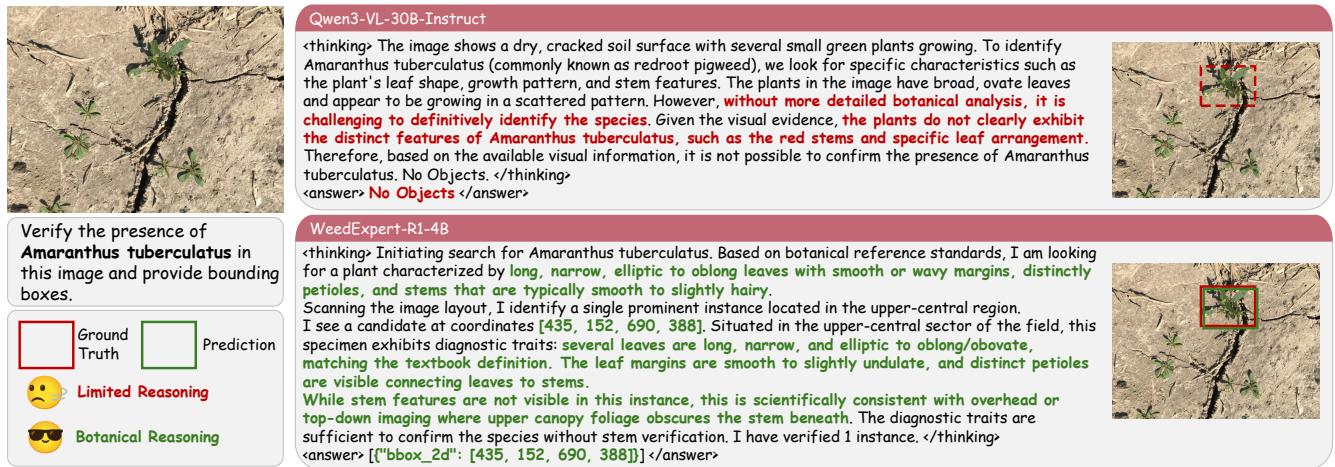}
    \caption{Comparison of the reasoning and grounding results produced by Qwen3-VL-30B-Instruct and our WeedExpert-R1-4B. Qwen3-VL-30B-Instruct fails to localize the target species due to \textcolor{myred}{insufficient botanical knowledge}, whereas the WeedExpert-R1-4B uses \textcolor{mygreen}{relevant morphological traits} and effectively \textcolor{mygreen}{distinguishes the target from visually similar plants and accurately localizes it}. All coordinates are scaled to [0, 1000].}\label{qwen-vs-ours}
    \vspace{-2em}
\end{figure*}

\begin{figure}[htbp]
  \centering
    \includegraphics[scale=0.5]{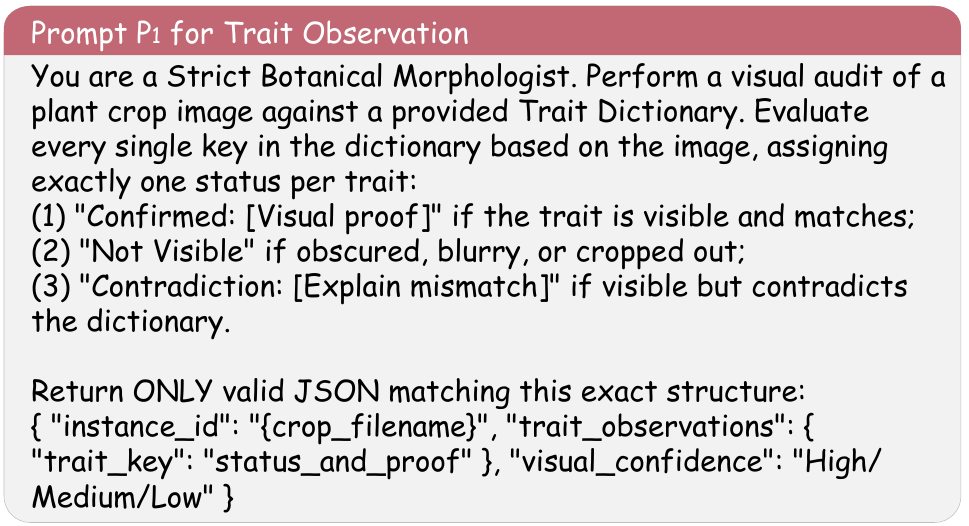}
    \caption{Prompt $P_{1}$ used  to generate trait observations (Eq.~\ref{gpt}). The Auditor evaluates whether each trait in the reference descriptor is visible in the cropped region of interest (RoI). Because the species identity is confirmed by the ground-truth annotation, the Auditor classifies each trait as confirmed, not visible, or contradictory. The resulting trait observations are subsequently used for CoT synthesis.}\label{prompt1}
\end{figure}

\begin{figure}[t]
  \centering
    \includegraphics[scale=0.5]{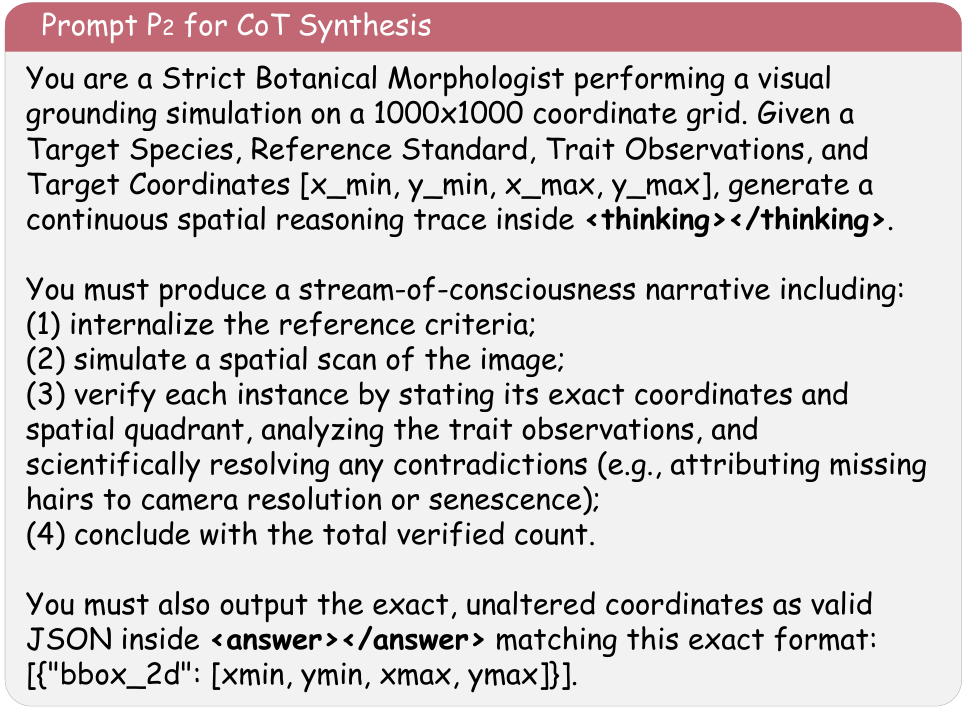}
    \caption{Prompt $P_{2}$ used for CoT synthesis (Eq.~\ref{gemini}). The synthesizer generates an image-level reasoning trace by using the aggregated trait observations, bounding-box annotations, and trait descriptor. No original image or cropped RoIs are provided at this stage, preventing the LLM from forming an independent global visual interpretation that could conflict with the per-instance trait observations from the visual-audit stage.}\label{prompt2}
\end{figure}

\subsection{Botanical Chain-of-Thought Synthesis}
\label{cot-data}
Weed grounding requires a model to identify and localize all instances of the targeted species specified in the prompt. Unlike general visual grounding, this task depends on understanding of key botanical traits, including leaf shape, leaf margin, petiole characteristics, leaf arrangement, and stem morphology. Although existing general-purpose or open-source MLLMs are pre-trained on large-scale human-curated corpora and may encode textual knowledge about many plant species, this knowledge is often shallow, and these models frequently fail to apply it consistently when distinguishing visually similar weeds. For example (Fig.~\ref{qwen-vs-ours}), while Qwen3-VL-30B-Instruct is a large, pre-trained open-source MLLM and produces a botanical-style explanation, it does not reliably associate the target species with the appropriate diagnostic traits or match those traits to the relevant plant in the image. Consequently, it fails to localize the target species. In contrast, our WeedExpert-R1-4B examines the visible morphological traits and uses them to identify and localize the target weed. This comparison suggests that the primary limitation of pre-trained MLLMs is not only access to textual botanical knowledge, but also the ability to align that knowledge with visual evidence during inference. To address this gap, a structured botanical reasoning process is introduced during the cold start stage to guide the model in examining diagnostic morphological traits before localizing the target plants.

A standard approach to synthesizing structured reasoning traces is to prompt advanced LLMs such as GPT~\citep{openai2026gpt54}, Gemini~\citep{deepmind2026gemini31}, and Claude~\citep{anthropic2026opus46} to generate CoT annotations directly from images. However, these annotations may contain imprecise botanical terminology or unsupported descriptions because general-purpose MLLMs do not consistently apply the specialized knowledge required for weed identification. Therefore, CoTs generated directly by general-purpose MLLMs may not provide sufficiently reliable supervision for domain-specific model training.

\begin{figure*}[t]
  \centering
    \includegraphics[scale=0.44]{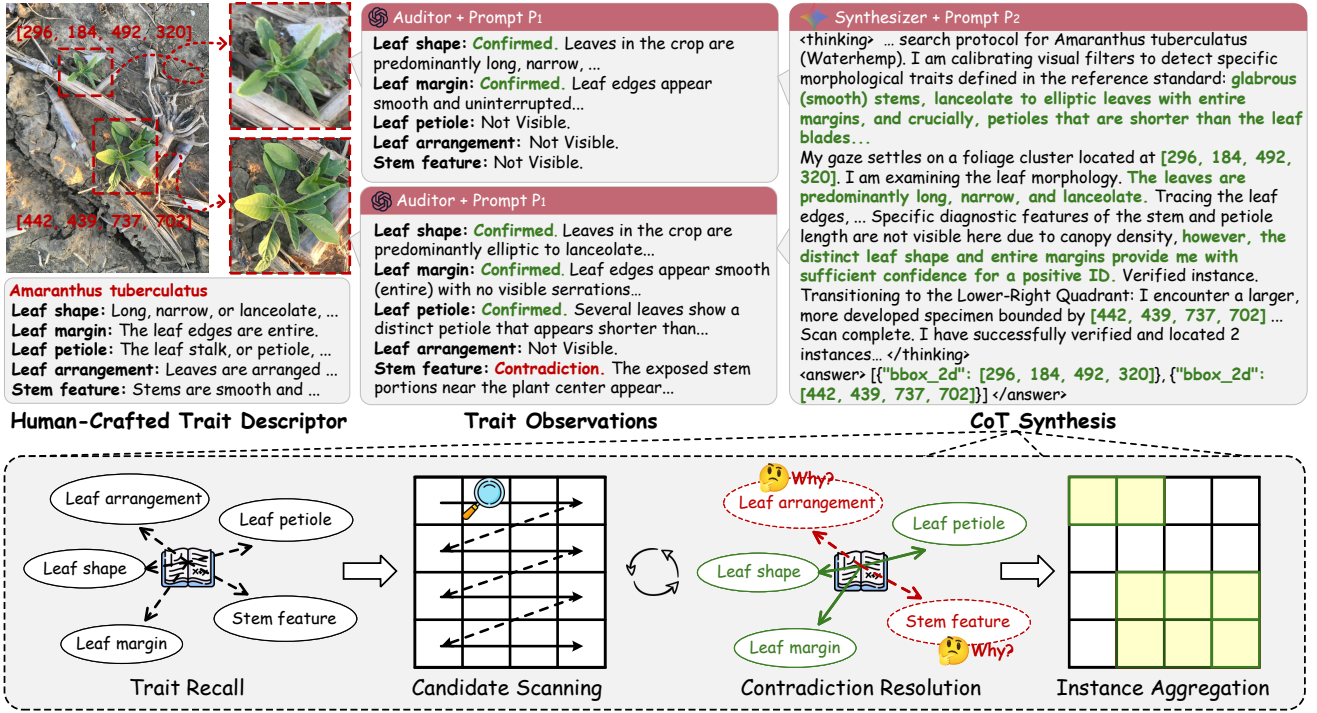}
    \caption{Workflow for CoT synthesis. In the first stage, GPT-5.4 takes Prompt $P_{1}$, the RoIs, and the corresponding trait descriptors to perform ''Visual Audit'', producing trait observations for each weed instance. In the second stage, Gemini-3.1-Pro takes Prompt $P_{2}$, the aggregated trait observations, the bounding box annotations, and the trait descriptors to synthesize the image-level domain-specific CoTs for each species. The bottom row summarizes the four-stage template followed by each synthesized CoT: Trait Recall (retrieving canonical traits), Candidate Scanning (sweeping regions that match for target traits), Contradiction Resolution (evaluating confirmed, occluded, or conflicting traits), and Instance Aggregation (collecting verified detections). Candidate scanning and contradiction resolution are repeated for each candidate instance.}\label{cot-syn}
    \vspace{-2em}
\end{figure*}

To improve the reliability of the synthesized CoTs, the proposed pipeline combines a human-crafted botanical trait dictionary with a visual auditing process: Given an image $I$ containing a target weed species $C$ with an instance $c$, together with its corresponding bounding box annotation $b$, two data inputs are prepared: (1) a region of interest (RoI), extracted via a $\textrm{Crop}(\cdot)$ function as $\textrm{Crop}(I, b)$; and (2) a morphological descriptor $D(C)$ retrieved from a human-crafted trait dictionary $D(\cdot)$, which records key botanical traits including leaf shape, leaf margin, leaf petiole, leaf arrangement, and stem feature. An auxiliary LLM (i.e., GPT-5.4), denoted $\textrm{Auditor}(\cdot)$, receives the instruction prompt $P_{1}$ (Fig.~\ref{prompt1}), the RoI, and the morphological descriptor as inputs and then performs a ``Visual Audit'' to determine whether each trait is visible in the RoI. The resulting trait observation (\textit{v}) is expressed as:
\begin{equation}
\label{gpt}
	v = \textrm{Auditor}(P_{1}, \textrm{Crop}(I, b), D(C))\ ,
\end{equation}
where the RoI constrains the auxiliary LLM attention to the target weed instance, thereby reducing interference from surrounding plants and limiting unsupported trait descriptions during visual verification.

A single image $I$ may contain multiple instances of the same weed species, each yielding its own RoI and a corresponding trait observation. For every target species $C$ present in $I$, the per-instance observations of a specific species are aggregated into a single CoT; an image containing $K$ distinct species therefore produces $K$ CoTs, one per (image, species) pair. For a given pair $(I, C)$ (image, species), aggregation yields an instance list $\textbf{c}=[c_{1}, c_{2}, \dots, c_{n}]$, the associated trait observations $\textbf{v}=[v_{1}, v_{2}, \dots, v_{n}]$, and the corresponding bounding box list $\textbf{b}=[b_{1}, b_{2}, \dots, b_{n}]$. Together with the morphological descriptor $D(C)$, $\textbf{v}$ and $\textbf{b}$ are passed to another auxiliary LLM, as a synthesizer, under the prompt $P_{2}$ (Fig.~\ref{prompt2}):

\begin{equation}
\label{gemini}
	\textrm{CoT} = \textrm{Synthesizer}(P_{2}, \textbf{v}, \textbf{b}, D(C))\ .
\end{equation}
During this stage, Gemini-3.1-Pro, used as the Synthesizer, is not provided with either the original image or the cropped RoIs (i.e., $\textbf{c}$). Instead, the synthesis is performed purely by comparing the textual trait observations against the descriptor, without any direct visual access to the image. This design prevents the LLM from forming a global visual interpretation that might contradict the per-instance observations obtained during the Visual Audit stage.

The synthesized CoT decomposes identification into four steps that resemble the diagnostic process of a \textit{Human Weed Scientist}. When a species is queried, the model first retrieves the target's morphological standards, including leaf shape, margin, arrangement, petiole, and stem feature, and states it explicitly as a reference standard before any pixels are examined (\textbf{Trait Recall}). The model then sweeps the image region by region, localizing each foliage cluster as a candidate and inspecting its locally observable traits against the reference standard (\textbf{Candidate Scanning}). When the observations are mixed (some traits confirmed, others occluded or inconsistent), the model reasons explicitly about the cause of each gap and decides whether the confirmed evidence alone is sufficient to commit to a positive identification (\textbf{Contradiction Resolution}). After all candidates has been evaluated, the verified instances are aggregated, and their bounding boxes are returned as the final answer.(\textbf{instance Aggregation}).



The complete CoT synthesis workflow is illustrated in Fig.~\ref{cot-syn}. Two auxiliary LLMs are employed in complementary roles across the two synthesis stages. The \emph{Auditor} (GPT-5.4) performs a visual audit of each RoI against the corresponding trait descriptor, producing the per-instance trait observations (Eq.~\ref{gpt}). The \emph{Synthesizer} (Gemini-3.1-Pro) then aggregates these observations into the final image-level CoT (Eq.~\ref{gemini}). Using two models from different model families increases the diversity of the synthesized CoT corpus and reduces the stylistic uniformity that may arise when all CoT are generated by a single model.

\begin{figure*}[!t]
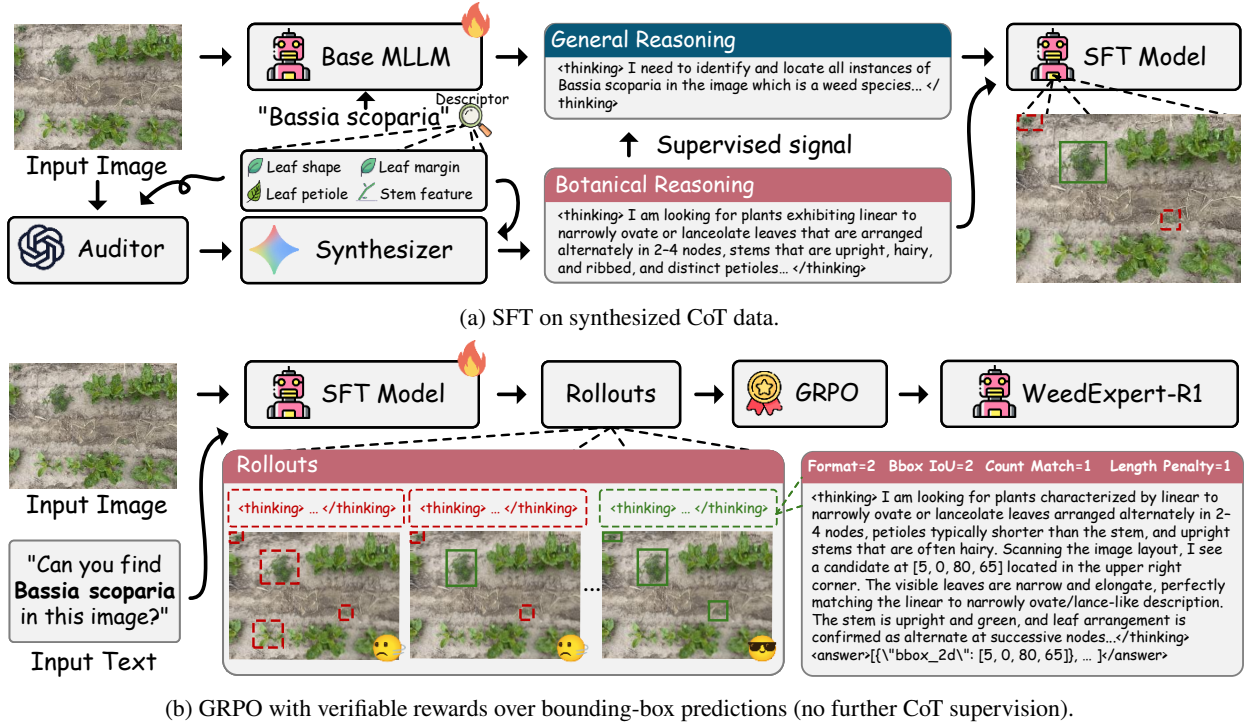

    \centering
    \begin{subfigure}[b]{0.95\linewidth}
        \centering
        \includegraphics[width=\linewidth,valign=t]{sft}
        \caption{SFT on synthesized CoT data.}
        \label{sft-frame}
    \end{subfigure}
    \begin{subfigure}[b]{0.95\linewidth}
        \centering
        \includegraphics[width=\linewidth,valign=t]{rl}
        \caption{GRPO with verifiable rewards over bounding-box predictions (no further CoT supervision).}
        \label{grpo-frame}
    \end{subfigure}
    \caption{Two-stage training framework of WeedExpert-R1. (a) In the SFT stage, the base MLLM is fine-tuned on synthesized botanical CoT data to establish structured botanical reasoning. (b) In the GRPO stage, the SFT model generates multiple rollouts for each image–species prompt and is optimized using composite verifiable rewards, including format, bbox IoU, count match, and length penalty rewards. This stage refines grounding precision without additional CoT supervision.}
    \label{workflow}
\end{figure*}

\subsection{Two-Stage Training of WeedExpert-R1}

After CoT synthesis, WeedExpert-R1 is trained in two stages (Fig~.\ref{workflow}). The first stage introduces supervised fine-tuning (SFT) as a cold start to establish structured botanical reasoning in an open-source MLLM. The second stage applies reinforcement learning (RL) to further refine the model's reasoning capability and improve grounding precision. Two separate datasets are constructed for the two stages.

The cold-start dataset $\mathcal{D}_{1} = (\mathcal{I}_{1}, \mathcal{C}_{1}, \mathcal{B}_{1})$ comprises images paired with their synthesized CoTs and corresponding bounding-box annotations. The aim of this stage is to format a well-structured botanical reasoning pattern that can be further refined during the subsequent RL stage. Because SFT provides direct supervision over the entire CoT and bounding-box sequence, it is particularly suited for shaping the model's reasoning structure. Training minimizes the negative log-likelihood over $\mathcal{D}_{1}$,
\begin{equation}
\mathcal{L}_{\text{SFT}} = -\mathbb{E}_{(I, C, B) \sim \mathcal{D}_{1}} \left[ \sum_{t=1}^{T} \log \pi_{\theta}(y_t \mid y_{<t}, I) \right]\ ,
\end{equation}
where $y=(C, B)$ is the target sequence comprising the CoT $C \in \mathcal{C}_{1}$ and $B\in \mathcal{B}_{1}$ denotes the bounding box annotations.

\begin{figure}[t]
  \centering
    \includegraphics[scale=0.51]{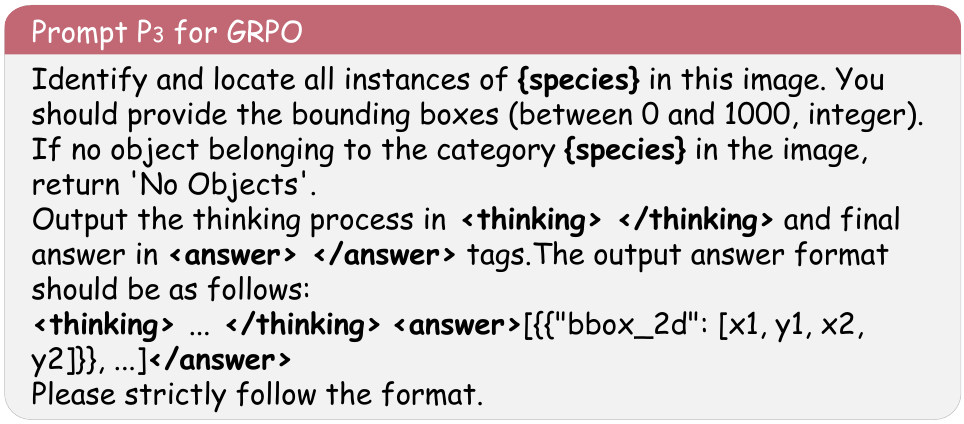}
    \caption{Prompt $P_{3}$ used for GRPO. The prompt instructs the model to produce a step-by-step reasoning trace within \texttt{<thinking>} and  \texttt{</thinking>} tags, followed by final bounding box predictions within \texttt{<answer>} and \texttt{</answer>} tags.}\label{prompt3}
    \vspace{-2em}
\end{figure}

The second stage applies GRPO with verifiable reward functions to refine grounding precision. No synthesized CoT data are used at this stage. The corresponding data set $\mathcal{D}_{2} = (\mathcal{I}_{2}, \mathcal{B}_{2})$ consists of images $I \in \mathcal{I}_{2}$ and their bounding box annotations $B \in \mathcal{B}_{2}$. During training, the model receives the prompt $P_{3}$ (Fig.~\ref{prompt3}), which instructs it to produce a step-by-step reasoning trace within \texttt{<thinking>} and \texttt{</thinking>} tags, followed by the final bounding box predictions within \texttt{<answer>} and \texttt{</answer>} tags. Four verifiable reward functions are defined to evaluate the model's responses.

\textbf{Format Reward $R_\textrm{format}$.} This reward enforces the required output structure and is composed of two components, each contributing 1. The first is awarded when the response contains a reasoning trace \texttt{<thinking>} and \texttt{</thinking>} tags. The second is awarded when the response contains an answer enclosed within \texttt{<answer>} and \texttt{</answer>} tags, and the answer is valid JSON containing the key \texttt{``bbox\_2d''}. The two components are summed, yielding $R_\textrm{format} \in \{0, 1, 2\}$.

\textbf{Bbox IoU Reward $R_{\textrm{bbox}}$.} This reward evaluates how well the predicted bounding boxes match the ground truth bounding boxes for a given image-species query pair $(I, C)$. The Hungarian algorithm~\citep{kuhn1955hungarian} is used to determine the optimal one-to-one matching between predictions and ground truth boxes. Each matched pair receives a discrete reward of $0$, $1$, or $2$ depending on whether it satisfies the two indicator functions: an IoU threshold $\tau_{\textrm{IoU}}$ and an $\tau_{L_{1}}$ distance threshold between centers of the prediction and ground truth boxes. The reward is computed as:

\begin{equation}
\begin{split}
R_{\textrm{bbox}} = \frac{1}{\max(n, k)} &\sum_{i=1}^{\min(n, k)} \biggl( \mathbb{1} \left[ \textrm{IoU}(b_i^{\textrm{gt}}, b_{\sigma(i)}^{\textrm{pred}}) > \tau_{\textrm{IoU}} \right] \\
&+ \mathbb{1} \left[ L_1(b_i^{\textrm{gt}}, b_{\sigma(i)}^{\textrm{pred}}) < \tau_{L_1} \right] \biggr)\ ,
\end{split}
\end{equation}
where $\sigma(\cdot)$ is the optimal assignment obtained using Hungarian algorithm, $\tau_{\textrm{IoU}}$ denotes the IoU threshold, set to $0.5$, and $\tau_{L_{1}}$ denotes the $L_{1}$ center-distance threshold, set to $10$ pixels. 

\textbf{Count Match Reward $R_{\textrm{count}}$.} This reward evaluates whether the model predicts the correct number of target instances for a given image–species query. It assigns a value of 1 when the number of predicted instances $n$ exactly matches the number of ground-truth instances $k$, and 0 otherwise. The reward is computed as:
\begin{equation}
	R_{\textrm{count}} = \mathbb{1}\left[ n=k \right]\ .
\end{equation}

\textbf{Length Penalty Reward $R_\textrm{length}$.} This reward penalizes excessively long reasoning. Let $l$ denote the length of the reasoning text enclosed within \texttt{<thinking>} and \texttt{</thinking>} tags and $L_{\textrm{max}}$ denote the maximum character budget, set to 4096 characters. The reward is 1 when the reasoning length does not exceed $L_{\textrm{max}}$. For long responses, the reward decreases linearly and is lower-bounded by $0$:
\begin{equation}
R_{\textrm{length}} = 
\begin{cases}  
1, & \text{if } l \leq L_{\max}, \\ 
\max\!\left(0,\; 1 - \dfrac{l - L_{\max}}{L_{\max}}\right), & \text{otherwise.} 
\end{cases}
\end{equation}

The final reward ($R$) is the sum of the four components:
\begin{equation}
	R = R_{\textrm{format}} + R_{\textrm{bbox}} + R_{\textrm{count}} + R_{\textrm{length}}\ .
\end{equation}

The total reward ranges from 0 to 6. The format and bbox IoU rewards each contribute up to 2 points, while the count match reward and length penalty reward each contribute up to 1 point.

\subsection{Experimental Setup}
\textbf{Benchmarks.} 
In this study, a new weed benchmark spanning 37 species was constructed from five publicly available weed datasets, namely 3SeasonWeedDet10~\citep{deng2025weed}, CottonWeedDet12~\citep{dang2023yoloweeds}, CottonWeedDet3~\citep{rahman2022deep}, Weed-crop~\citep{upadhyay2025weed}, and Weed25~\citep{wang2022weed25}, together with a dataset collected at the Panhandle Research and Extension Center (PREEC) (Scottsbluff, NE, USA). The combined benchmark spans a broad geographic range, with detailed statistics reported in Table~\ref{benchmarks}. Because 3SeasonWeedDet10 and CottonWeedDet12 share a subset of images, duplicated annotations were removed. 

The combined dataset was first partitioned into an SFT cold-start portion (5\%) and an RL portion (95\%). The SFT subset was constructed using stratified sampling across each source dataset–species pair. To ensure broad coverage, at least one example was first selected from each pair, so that all species and the visual patterns of all six source datasets were represented. The remaining SFT sampling budget was then allocated to each pair in proportion to its size, allowing common species and source datasets to retain their natural representation while preventing rare species from being omitted. This strategy ensured that the synthesized CoT reasoning traces covered the species diversity and visual variability present during training. The RL portion was further divided into training, validation, and testing sets, accounting for 75\%, 10\%, and 10\%. The SFT portion was used to synthesize CoT annotations (Sec.~\ref {cot-data}), while the RL portion contained only image–species queries and their corresponding bounding-box annotations.
  
\begin{table*}[!htp]
\centering
\small
\caption{Composition of the benchmark datasets used in this study. The benchmark was constructed from five publicly available weed datasets and one dataset collected at the University of Nebraska–Lincoln Panhandle Research and Extension Center (PREEC). Weed25$^{\dag}$ was sourced from Roboflow~\citep{roboflow} as the original link is no longer available. For images containing multiple weed species, separate image-species entries were generated; for example, an image containing two species $A$ and $B$ yields two entries, $\langle\text{image}, \mathbf{b}_{A}\rangle$ and $\langle\text{image}, \mathbf{b}_{B}\rangle$. The source datasets contain 56$^{\ddag}$ species labels in total, corresponding to 37 unique species after duplicate species across datasets were merged.}\label{benchmarks}
\begin{tabular*}{\tblwidth}{@{}lccccc@{}}
\toprule
Dataset & Images & Weed Species & Resolution & Region & Final Entries \\
\midrule
3SeasonWeedDet10~\citep{deng2025weed}    & 8,463 & 10 & $\sim$3K, $\sim$4K & Southern US        & 9,743 \\
CottonWeedDet12~\citep{dang2023yoloweeds}& 5,648 & 12 & $\sim$4K & Southern US        & 922   \\
CottonWeedDet3~\citep{rahman2022deep}    &   848 &  3 & $\sim$4K, $\sim$8K & Southern US        & 923   \\
Weed-crop~\citep{upadhyay2025weed}       & 1,120 &  5 & $\sim$6K, $\sim$8K & North Dakota, US   & 1,816 \\
Weed25$^{\dag}$~\citep{wang2022weed25}   & 9,553 & 25 & $640\times 640$ & Chongqing, China   & 9,553 \\
PREEC (Ours)                             &   202 &  1 & $\sim$4K & Nebraska, US       & 202   \\
\midrule
\textbf{Total}                           & \textbf{25,834} & \textbf{37 (56$^{\ddag}$)} & --- & --- & \textbf{23,159} \\
\bottomrule
\end{tabular*}
\end{table*}

    \begin{figure*}[t]
    \centering
    \begin{subfigure}[t]{0.42\linewidth}
        \centering
        \includegraphics[width=\linewidth,valign=t]{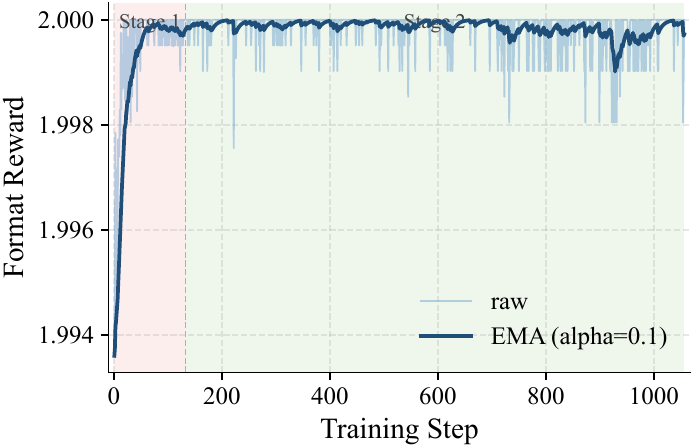}
        \caption{Format Reward $R_\textrm{format}$ over 1050 steps.}
        \label{reward-format}
    \end{subfigure}
    \hspace{2em}
    \begin{subfigure}[t]{0.42\linewidth}
        \centering
        \includegraphics[width=\linewidth,valign=t]{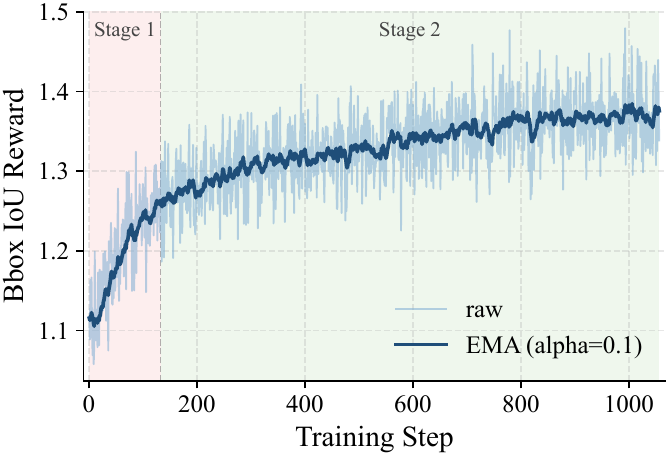}
        \caption{Bbox IoU Reward $R_{\textrm{bbox}}$ over 1050 steps.}
        \label{reward-accuracy}
    \end{subfigure}
     \begin{subfigure}[b]{0.42\linewidth}
        \centering
        \includegraphics[width=\linewidth,valign=t]{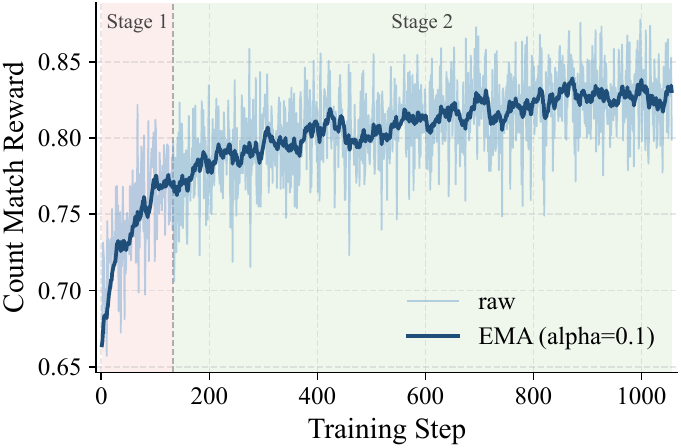}
        \caption{Count Match Reward $R_{\textrm{count}}$ over 1050 steps.}
        \label{reward-count}
    \end{subfigure}
    \hspace{2em}
     \begin{subfigure}[b]{0.42\linewidth}
        \centering
        \includegraphics[width=\linewidth,valign=t]{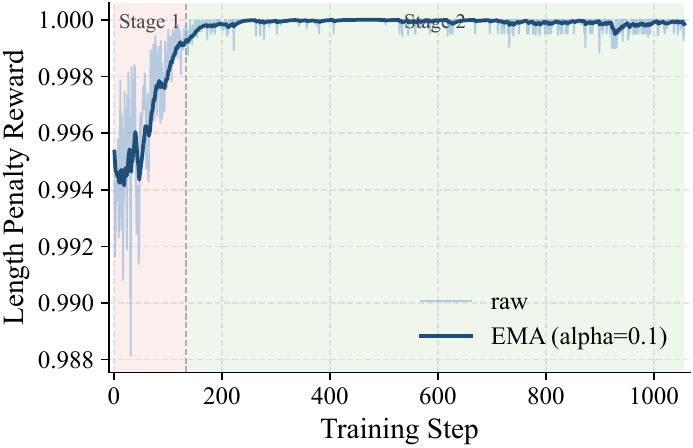}
        \caption{Length Penalty Reward $R_\textrm{length}$ over 1050 steps.}
        \label{reward-length-penalty}
    \end{subfigure}
    \caption{GRPO training dynamics across four verifiable rewards over 1050 training steps. Step 0 represents the initial reward values achieved by the SFT checkpoint before GRPO optimization. Stage 2 indicates the later optimization period, when grounding-related rewards continue to improve more gradually. Curves are smoothed using an exponential moving average (EMA, $\alpha=0.1$) to better visualize overall trends}.
    \label{dynamics}
    \vspace{-2em}
\end{figure*}
  
 \textbf{Baselines.}
 WeedExpert-R1 was evaluated against six baselines:
 Two advanced proprietary models, including GPT-5.4~\citep{openai2026gpt54}, Gemini-3.1-Pro~\citep{deepmind2026gemini31}, and four state-of-the-art open-source models, including Qwen3-VL-4B-Instruct, Qwen3-VL-30B-Instruct~\citep{bai2025qwen3}, Gemma-4-E4B-it, Gemma-4-31B-it~\citep{google2026gemma4}. All baseline models were evaluated without task-specific fine-tuning using the same weed-grounding prompt.
  
 \textbf{Training Details.}
 The base model for WeedExpert-R1, Qwen3-VL-4B-Instruct, was first fine-tuned using SFT on domain-specific botanical CoT annotations to establish a structured reasoning pattern, and subsequently trained with GRPO to refine grounding precision. For the SFT stage, training was conducted for 5 epochs on 4$\times$ H100 GPUs, with a global batch size of 96 and a maximum sequence length of 8192 tokens. The GRPO stage was initialized from the SFT checkpoint and trained for 1050 steps with a global batch size of 128. For each prompt, $n=8$ rollouts were sampled, with response length capped at 4096 tokens and prompt length capped at 6144 tokens. The KL coefficient was set to $0$ to encourage policy exploration. Due to computational constraints, images whose longer side exceeded 2K resolution were downsampled to 2K.
  
 \textbf{Metrics.}
 Precision weed grounding follows the task formulation and evaluation framework of Generalized Referring Expression Comprehension (GREC)~\citep{he2023grec}, in which the model must localize all object instances referred to by a natural language prompt. Conventional detectors typically report mean Average Precision (mAP), which relies on per-prediction confidence scores to compute precision–recall curves. Because MLLM outputs do not provide such confidence scores, mAP cannot be applied directly. Instead, Precision@0.5 and Recall@0.5 were reported, where 0.5 denotes the IoU threshold above which a prediction is considered correct. In the single-instance case, Precision@0.5 reduces to top-1 accuracy. Because each image may contain multiple weed instances, the micro-averaged metrics, mPrecision@0.5 and mRecall@0.5, were also reported. The micro-averages were computed at the bounding box level rather than the image level, treating each ground-truth instance as an independent unit. Finally, a stricter metric, Precision@(F$_1$=1, IoU$\geq$0.5)~\citep{he2023grec}, was adopted. The metric is defined as the proportion of image-species queries for which the model achieves F$_1$=1,  meaning every ground-truth instance is correctly localized with no false positives and no false negatives. All evaluations used greedy decoding with temperature set to $0$ to ensure deterministic text output.
 
\textbf{Experiments.}
The experimental evaluation consisted of three parts. First, the performance gain of each training stage was examined by comparing four checkpoints: the base Qwen3-VL-4B-Instruct, the SFT-only model, the GRPO-only model, and the full WeedExpert-R1, trained with SFT followed by GRPO. Both quantitative metrics and qualitative response analysis were reported. Second, WeedExpert-R1 was compared with all six baseline models: two proprietary frontier models and four state-of-the-art open-source MLLMs. Third, model's robustness and generalization were evaluated through two qualitative studies: one using challenging test images that require fine-grained botanical reasoning to distinguish visually similar instances, and another using an open-vocabulary evaluation on weed and crop species absent from the post-training data.


\section{Results and Discussion}
\label{main-results}

\textbf{Training Dynamics.}
The GRPO training dynamics (shown in Fig.~\ref{dynamics}), revealed two empirical stages of reward improvement. Stage 1 corresponds to the early optimization period, during which the model rapidly satisfies the structural requirements of the training. The format reward $R_{\textrm{format}}$ and the length penalty reward $R_\textrm{length}$ rapidly reached their maximum values of $2.0$ and $1.0$, respectively. This indicates that the SFT cold-start stage had already established the required response format and length control before GRPO optimization. During stage 1, the SFT-initialized model already produced well-formatted responses, allowing the grounding-related rewards, including the bounding-box IoU reward $R_{\textrm{bbox}}$ and count match reward $R_{\textrm{count}}$ to increase during GRPO exploration. Stage 2 corresponds to the later optimization period, during which the structural rewards remain stable, and further improvement is mainly driven by the harder grounding-related objectives in the training set. Specifically, $R_{\textrm{bbox}}$ steadily improved from $\sim$1.28 to $\sim$1.40 and $R_{\textrm{count}}$ increased from approximately $\sim$0.76 to $\sim$0.83. These dynamics indicate that the GRPO training follows a natural, easy-to-hard progression: the model first secures structural compliance and then refines its grounding accuracy on more challenging cases to improve the bbox IoU and count match rewards.


 \begin{figure*}[!t]
  \centering
    \includegraphics[scale=0.82]{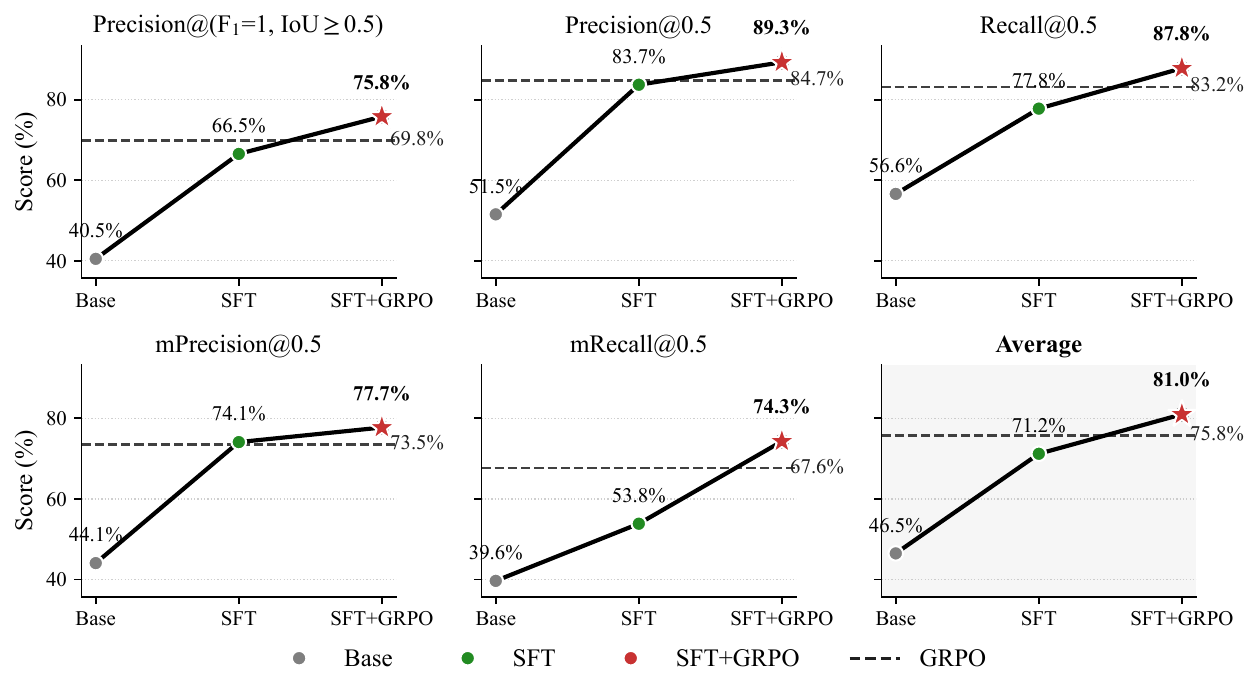}
    \caption{Performance gains across the five evaluation metrics for Qwen3-VL-4B-Instruct (\textbf{Base}), cold-start SFT-only (\textbf{SFT}), and the full WeedExpert-R1 model trained with SFT followed by GRPO(\textbf{SFT + GRPO}). The dashed line in each panel represents direct GRPO training applied to the base model without the SFT stage (\textbf{GRPO}). The \textit{Average} panel reports the mean across the five metrics. \textcolor{myred}{Stars indicate the best-performing training strategy in each panel}. The SFT+GRPO pipeline achieved the highest score on individual metrics and on the overall average.}\label{gains}
    \vspace{-1em}
\end{figure*}

\textbf{Effect of SFT and GRPO on Weed Grounding Performance.}
The contribution of each training stage was evaluated by comparing four model variants: the base model (Qwen3-VL-4B-Instruct), the SFT-only model, the GRPO-only model (direct GRPO trained without the cold-start stage), and the full WeedExpert-R1 model (trained with 
SFT followed by GRPO) (Fig.~\ref{gains}). Four key patterns were observed from the comparison. First, the base model performed poorly on species-level weed grounding, which is a limitation of a general-purpose MLLM without specialized botanical domain knowledge. Second, GRPO without the SFT improved performance over the base model, indicating that the verifiable reward signals provided useful supervision. However, compared with SFT alone, the gains were marginal and inconsistent across metrics; for example, GRPO was 0.6$\%$ lower than SFT on mPrecision@0.5. This suggests that GRPO alone can improve grounding but is less effective than GRPO initialized from a structured botanical reasoning model. Third, SFT alone substantially improved performance over the base model by introducing botanical CoT supervision and establishing a structural botanical reasoning framework. Fourth, applying GRPO after the SFT checkpoint further improved the model's grounding capability across all five performance metrics. 

The trend was most evident in the strict Precision@($\textrm{F}_1{=}1$, $\text{IoU}{\geq}0.5$) metric. The base model achieved only $40.5\%$ on Precision@($\textrm{F}_1{=}1$, $\text{IoU}{\geq}0.5$), whereas SFT increased the score to $66.5\%$. GRPO further improved the score to $69.8\%$, while the full SFT + GRPO pipeline achieved the highest score of $75.8\%$. This represents an improvement of $+35.3$ points over the base model, $+9.3$ points over SFT, and $+6.0$ points over GRPO. Similar trends hold across the other four metrics, with SFT + GRPO achieving the best performance score on every metric and increasing the average score from $46.5\%$ for the base model to $81.0\%$. The largest additional gain from SFT + GRPO was observed for mRecall@0.5, which increased from $53.8\%$ to $74.3\%$. This improvement suggests GRPO directly optimizes verifiable grounding rewards, including bounding-box IoU and instance-count matching, which encourage the model to localize a more complete set of target instances in multi-instance scenes.

\begin{figure*}[!t]
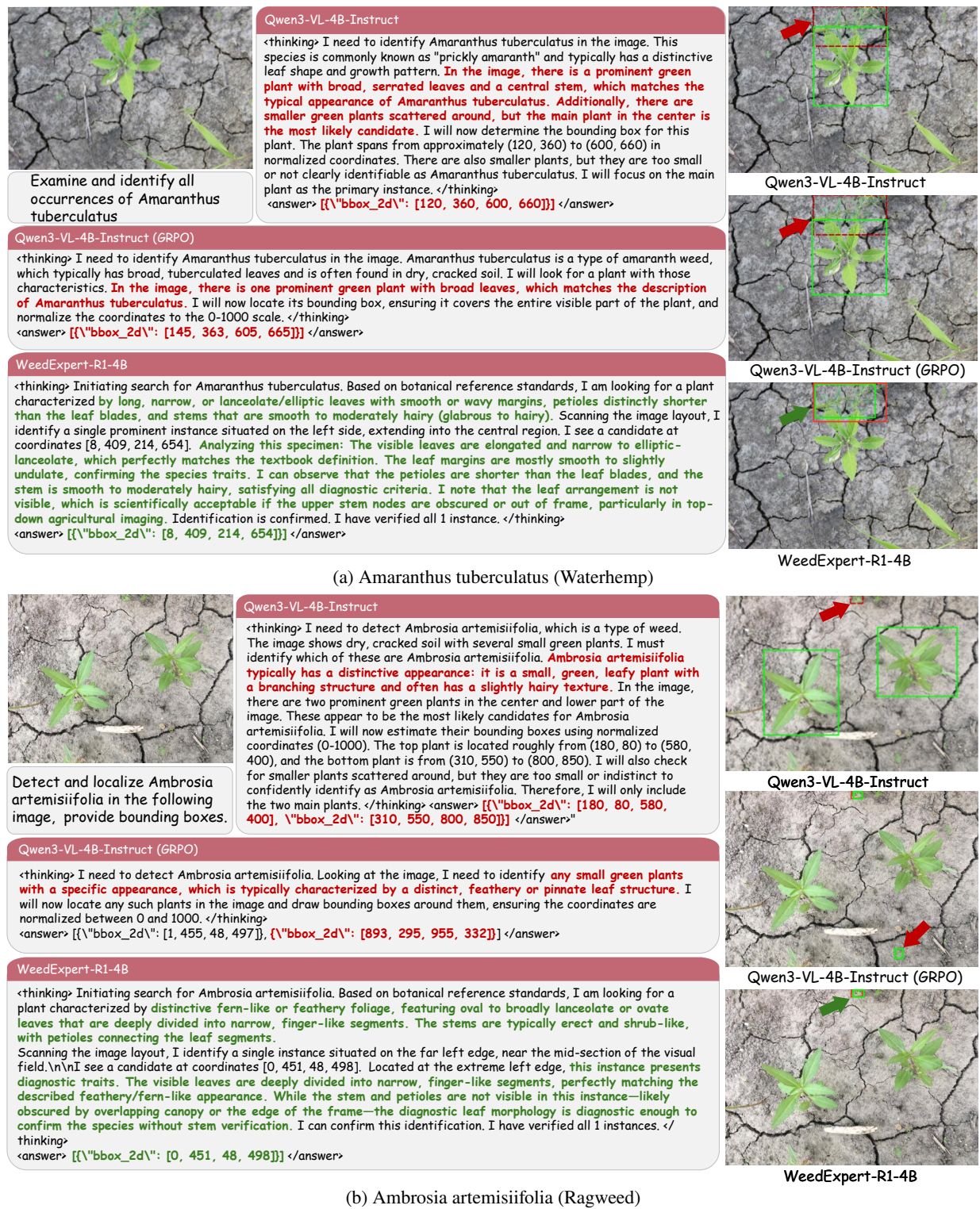

    \centering
    \begin{subfigure}[b]{0.94\linewidth}
        \centering
        \includegraphics[width=\linewidth]{gains_fig_a}
        \vspace{-2em}
        \caption{Amaranthus tuberculatus (Waterhemp)}
        \label{comp1}
    \end{subfigure}
    \begin{subfigure}[b]{0.94\linewidth}
        \centering
        \includegraphics[width=\linewidth]{gains_fig_b}
        \vspace{-2em}
        \caption{Ambrosia artemisiifolia (Ragweed)}
        \label{comp2}
    \end{subfigure}
    \caption{Comparison of model responses and grounding results from the Qwen3-VL-4B, the GRPO model without SFT, and WeedExpert-R1-4B trained with SFT + GRPO. \textcolor{myred}{Red boxes} denote ground truth annotation; \textcolor{mygreen}{green boxes} denote model predicted grounding box. Both examples were rotated 90$^\circ$ clockwise for display.}
    \label{fig_gains}
\end{figure*}

The reasoning behavior of the Qwen3-VL-4B-Instruct, GRPO, and SFT + GRPO (WeedExpert-R1-4B) was compared using two challenge test images, each containing multiple weed species that could lead to visual confusion (Fig.~\ref{fig_gains}). Compared with the relatively straightforward example shown in Fig.~\ref{qwen-vs-ours}, these cases are more difficult because the target species are small, partially occluded, or surrounded by visually similar broadleaf weeds. The base model generated relatively long reasoning traces but its descriptions focus on generic visual features rather than botanical traits, resulting in incorrect grounding. For example, in the \textit{Amaranthus tuberculatus} case, the base model emphasized broad, serrated leaves and selected a visually prominent plant, whereas WeedExpert-R1 identified the target using more specific traits, including long, narrow to lanceolate leaves, smooth to slightly wavy margins, shorter petioles, and stem characteristics (Fig.~\ref{fig_gains}(a)). Similarly, in the \textit{Ambrosia artemisiifolia} case, the base model selected larger and more visually obvious plants, while WeedExpert-R1 localized the small target instance by relying on deeply divided, fern-like leaf morphology (Fig.~\ref{fig_gains}(b)).

GRPO produced some correct predictions, but its reasoning traces were markedly shorter and less informative than those of the base model (Fig.~\ref{fig_gains}). This degenerate behavior is a case of \textbf{reward hacking}~\citep{shen2025vlm, fu2025reward}, a known failure mode in RL, in which the policy exploits the reward signal by producing minimal reasoning rather than learning the intended diagnostic process. As a result, the final grounding performance of GRPO remained limited. In contrast, the SFT + GRPO pipeline mitigated the failure by first establishing a structured botanical reasoning pattern through cold start on Auditor–Synthesizer-generated CoT samples and then refining grounding precision through GRPO. Overall, WeedExpert-R1 systematically compared visible traits with the target species’ botanical profile, evaluated missing or occluded trait evidence, and localized the target instances more accurately. These qualitative results support the quantitative findings by showing that the Auditor–Synthesizer CoT supervision provides a structured reasoning foundation that enables GRPO to improve grounding accuracy while maintaining botanical interpretability.

\newcommand{\nogain}{\phantom{\textcolor{ForestGreen}{\scriptsize$_{+0.0000}$}}}
\begin{table*}[!t]
\centering
\small
\caption{Performance comparison of WeedExpert-R1 against six baseline MLLMs on species-level weed grounding. Because each image may contain multiple weed instances, Precision@0.5 and Recall@0.5 are computed at both the image level (macro) and bounding-box level (micro). The best results in each column are shown in \textbf{bold}. \textcolor{ForestGreen}{Green} subscripts indicate the absolute improvements over the Qwen3-VL-4B-Instruct, the base model used for WeedExpert-R1.}
\label{main_results}
\setlength{\tabcolsep}{3pt}
\begin{tabular}{lcccccc}
\toprule
\multirow{2}{*}{\textbf{Model}} & \multirow{2}{*}{\textbf{Precision@(F$_{1}$=1, IoU$\geq$0.5)}} & \multicolumn{2}{c}{\textbf{Macro (image-wise)}} & & \multicolumn{2}{c}{\textbf{Micro (bbox-wise)}} \\
\cmidrule(lr){3-4} \cmidrule(lr){6-7}
 & & Precision@0.5 & Recall@0.5 & & mPrecision@0.5 & mRecall@0.5 \\
\midrule
\rowcolor{gray!20}
\multicolumn{7}{l}{\textbf{\textit{Proprietary models}}} \\
GPT-5.4         & 0.4600\,\nogain & 0.5303\,\nogain & 0.5392\,\nogain & & 0.4012\,\nogain & 0.3261\,\nogain \\
Gemini-3.1-Pro  & 0.6158\,\nogain & 0.7548\,\nogain & 0.7693\,\nogain & & 0.6506\,\nogain & 0.5796\,\nogain \\
\midrule
\rowcolor{gray!20}
\multicolumn{7}{l}{\textbf{\textit{Open-source models}}} \\
Gemma-4-E4B-it       & 0.3027\,\nogain & 0.3819\,\nogain & 0.4190\,\nogain & & 0.3333\,\nogain & 0.2624\,\nogain \\
Qwen3-VL-4B-Instruct & 0.4050\,\nogain & 0.5154\,\nogain & 0.5662\,\nogain & & 0.4406\,\nogain & 0.3965\,\nogain \\
Gemma-4-31B-it       & 0.4455\,\nogain & 0.5460\,\nogain & 0.6052\,\nogain & & 0.3433\,\nogain & 0.4019\,\nogain \\
Qwen3-VL-30B-Instruct& 0.4795\,\nogain & 0.5809\,\nogain & 0.5748\,\nogain & & 0.6006\,\nogain & 0.4136\,\nogain \\
\hdashline
\rowcolor{gray!20}
\multicolumn{7}{l}{\textbf{\textit{Ours (Qwen3-VL-4B-Instruct)}}} \\
\quad + SFT
  & 0.6655\,\textcolor{ForestGreen}{\scriptsize$_{+0.2605}$}
  & 0.8372\,\textcolor{ForestGreen}{\scriptsize$_{+0.3218}$}
  & 0.7779\,\textcolor{ForestGreen}{\scriptsize$_{+0.2117}$}
  & & 0.7410\,\textcolor{ForestGreen}{\scriptsize$_{+0.3004}$}
  & 0.5384\,\textcolor{ForestGreen}{\scriptsize$_{+0.1419}$} \\
\quad + GRPO
  & 0.6982\,\textcolor{ForestGreen}{\scriptsize$_{+0.2932}$}
  & 0.8472\,\textcolor{ForestGreen}{\scriptsize$_{+0.3318}$}
  & 0.8318\,\textcolor{ForestGreen}{\scriptsize$_{+0.2656}$}
  & & 0.7354\,\textcolor{ForestGreen}{\scriptsize$_{+0.2948}$}
  & 0.6762\,\textcolor{ForestGreen}{\scriptsize$_{+0.2797}$} \\
\quad + SFT + GRPO (WeedExpert-R1)
  & \textbf{0.7582}\,\textcolor{ForestGreen}{\scriptsize$\boldsymbol{_{+0.3532}}$}
  & \textbf{0.8930}\,\textcolor{ForestGreen}{\scriptsize$\boldsymbol{_{+0.3776}}$}
  & \textbf{0.8781}\,\textcolor{ForestGreen}{\scriptsize$\boldsymbol{_{+0.3119}}$}
  & & \textbf{0.7772}\,\textcolor{ForestGreen}{\scriptsize$\boldsymbol{_{+0.3366}}$}
  & \textbf{0.7428}\,\textcolor{ForestGreen}{\scriptsize$\boldsymbol{_{+0.3463}}$} \\
\bottomrule
\end{tabular}
\end{table*}

\begin{figure*}[t]
  \centering
    \includegraphics[scale=0.43]{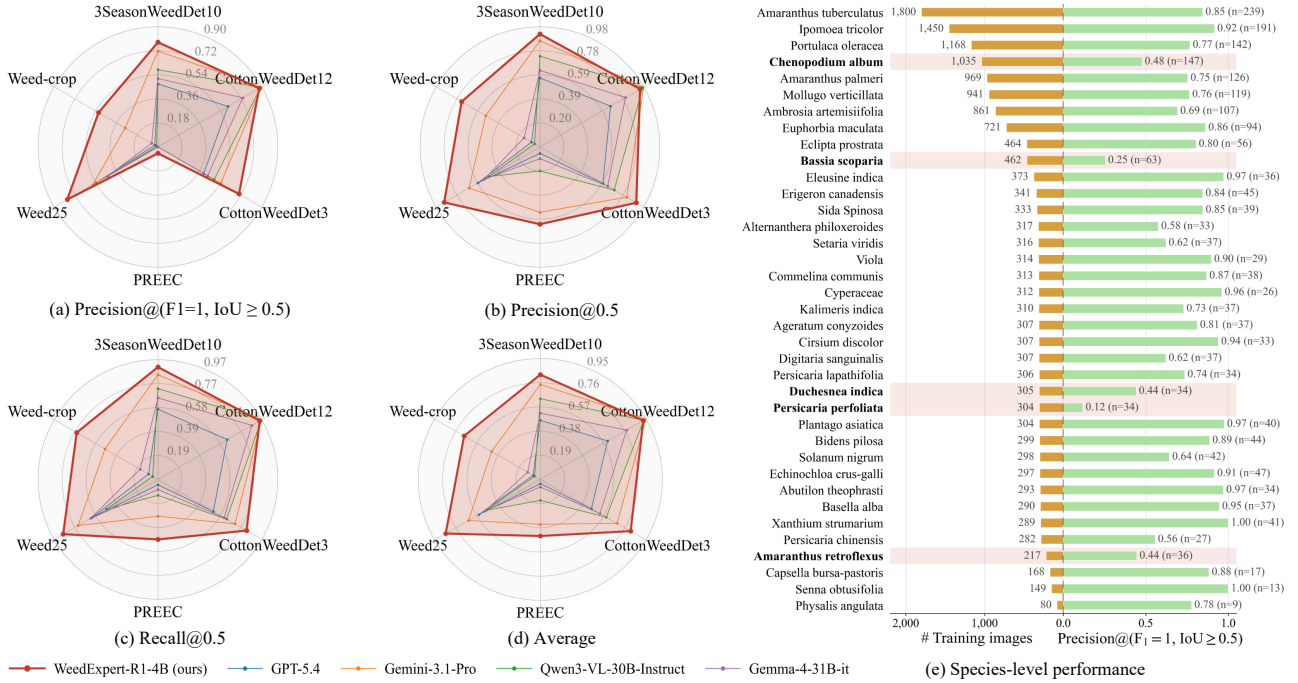}
    \caption{Dataset-level and species-level performance of WeedExpert-R1 and baseline MLLMs on the weed-grounding benchmark. (a)--(d) Performance metrics across source datasets for Precision@($\text{F}_1=1$, IoU$\geq$0.5), Precision@0.5, Recall@0.5 and the average of the three metrics. The ``PREEC'' dataset, which contains in-field images with mixed weed and crop instances, presents a challenging scenario for all models. (e) Species-level performance of WeedExpert-R1 compared with the number of training images per species. Orange bars indicate the number of training dataset images (left), and green bars indicate Precision@($\text{F}_1=1$, IoU$\geq$0.5), with the test-set size \textit{n} shown in parentheses. Species with low grounding performance, defined as a score $\leq 0.5$, are highlighted.}\label{chart}
    \vspace{-2em}
\end{figure*}

\textbf{Comparison with Frontier Models and Baselines}
Performance across the five evaluation metrics is reported in Table~\ref{main_results}. All models were prompted using the same instruction to output bounding boxes normalized to $[0,1000]$, and no model-specific filtering or correction was applied to invalid bounding-box outputs to ensure a fair comparison. Even frontier models that demonstrate strong general visual-language capabilities, such as GPT-5.4 and Gemini-3.1-Pro, struggled with precise species-level weed grounding. Among the open-source baselines, larger models generally perform better: Qwen3-VL-30B-Instruct and Gemma-4-31B-it outperforming their 4B and E4B counterparts, respectively. However, a substantial performance gap remained, with all five metrics for the baseline models falling within approximately $50\% \sim 60\%$.

With only $5\%$ of the total data used for CoT annotation synthesis in the SFT stage, WeedExpert-R1-4B substantially outperformed the Qwen3-VL-30B-Instruct baseline, achieving a $+26.05$ gain on Precision@($\textrm{F}_1{=}1$, $\text{IoU}{\geq}0.5$), and an average gain of approximately $+25$ points across all five metrics. The results highlight the critical role of the botanical reasoning trace generated by the Auditor–Synthesizer pipeline. GRPO, which bypassed the SFT stage, increased Precision@($\textrm{F}_1{=}1$, $\text{IoU}{\geq}0.5$) to only $69.82\%$, and produced less informative reasoning traces. By contrast, applying GRPO after SFT increased the metric to $75.82\%$, representing a $+35.32$ point gain over the base model, while preserving reasoning grounded in botanical domain knowledge. Overall, WeedExpert-R1-4B achieved an average gain of approximately $+35$ points across the five metrics and surpassed both frontier proprietary models and state-of-the-art open-source baselines. These findings suggest that domain-specific botanical CoT synthesis, rather than model scale alone, is essential for reliable weed grounding.

Fig.~\ref{chart}(a)--(d) compares WeedExpert-R1 with four baselines, including GPT-5.4, Gemini-3.1-Pro, Qwen3-VL-30B-Instruct, and Gemma-4-31B-it) on the combined weed benchmark. The comparison includes Precision@($\textrm{F}_1{=}1$, $\text{IoU}{\geq}0.5$), Macro Precision@0.5, Recall@0.5, and the average across these metrics. All models performed best on CottonWeedDet12, where most images contain only a single weed instance, and therefore present fewer localization ambiguities. In contrast, PREEC was the most challenging dataset. Although PREEC contains only one labeled target, kochia (\textit{Bassia scoparia}), all images were captured in real agricultural production fields with crop plants and other unlabeled weed species present, creating complex multi-instance scenes. Despite these challenges, WeedExpert-R1-4B consistently outperformed all other frontier proprietary models and larger open-source baselines across the benchmark datasets. These results suggest that the Auditor–Synthesizer-based botanical CoT helps the model transfer weed-science trait knowledge into more reliable visual grounding, rather than relying on model scale alone.

\begin{figure*}[!t]
  \centering
    \includegraphics[scale=0.52]{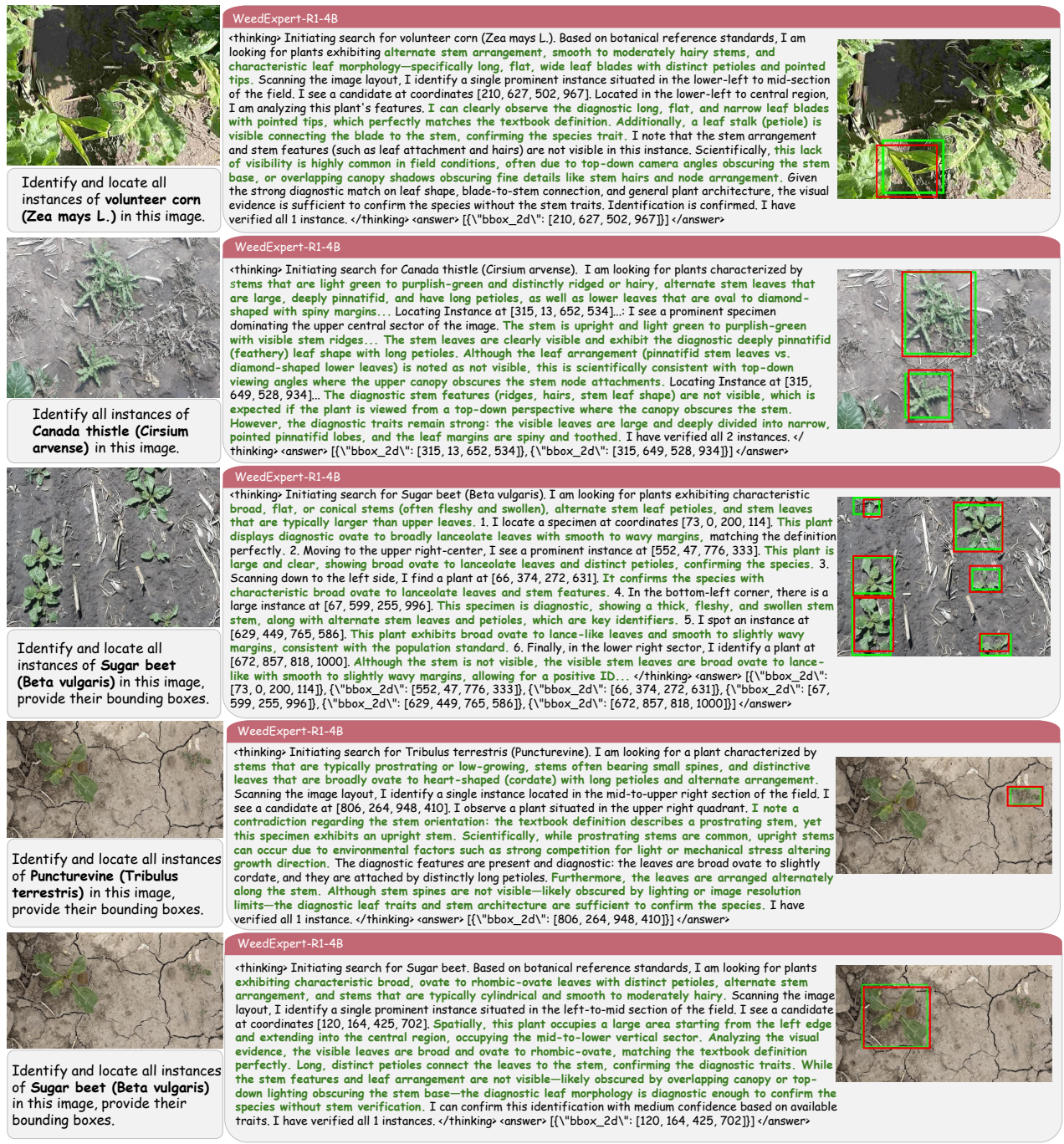}
    \caption{Qualitative examples of open-vocabulary weed grounding by WeedExpert-R1-4B. The three weed species (\textit{Cirsium arvense}, \textit{Zea mays}, \textit{Tribulus terrestris}) and one crop (\textit{Beta vulgaris}) shown here were absent from the SFT and GRPO post-training data. WeedExpert-R1-4B localizes the prompted target instances in these examples by using botanical trait-based reasoning. \textcolor{myred}{Red} boxes indicate ground-truth annotations, and \textcolor{mygreen}{green} boxes indicate model predictions.}\label{open}
    \vspace{-2em}
\end{figure*}

\textbf{Species-Level Grounding Performance.} Fig.~\ref{chart}(e) reports per-species Precision@($\textrm{F}_1{=}1$, $\text{IoU}{\geq}0.5$) with the number of training images for each species. Species with scores below 0.5 were highlighted in red. Lower performance for these species may reflect limited training instances, less distinctive or highly variable visual traits, and challenging detection scenarios such as high instance density or strong visual similarity among species. For example, \textit{Chenopodium album} images in 3SeasonWeedDet10 often contain high target-instance counts, with more than 20 plants in a single image, making complete instance-level localization more challenging. Similarly, many \textit{Bassia scoparia} images in the PREEC dataset were collected under real agricultural production conditions, where the target plants appear alongside crops and other unlabeled weed species. \textit{Amaranthus retroflexus} also showed lower precision, likely because it can be visually confused with related or morphologically similar species such as \textit{Amaranthus palmeri} and \textit{Chenopodium album}. In contrast, our WeedExpert-R1 achieved high precision for several species with distinctive botanical traits, even when the number of training images was relatively small, $\leq 300$. This pattern suggests that species-level grounding performance is influenced not only by training image count, but also by the visibility and distinctiveness of diagnostic traits. It further supports the value of the Auditor–Synthesizer pipeline, which converts expert botanical trait knowledge into structured reasoning traces for weed grounding.

\textbf{Open-Vocabulary Weed Grounding.} Weed species vary substantially across regions, crops, and production systems, and the species requiring management may differ from one field to another. A model capable of grounding diverse weed species, including species unseen during post-training, could be adapted more easily across fields, crops, and geographic regions without redefining a fixed detector class list, making open-vocabulary weed grounding an important research direction despite its inherent difficulty. 
To qualitatively demonstrate the open-vocabulary capability of WeedExpert-R1, four representative scenarios were tested (Fig.~\ref{open}): three target plants absent from the SFT and GRPO post-training data, including (\textit{Cirsium arvense}, \textit{Zea mays}, and \textit{Tribulus terrestris}) (Table~\ref{benchmarks}), and one crop species \textit{Beta vulgaris} (sugar beet). These examples cover challenging field conditions, including top-down field imagery, partial occlusion, and multiple co-occurring plant instances. Across these selected examples, WeedExpert-R1-4B localized the prompted target species and generated reasoning traces based on species-specific botanical traits. These qualitative results suggest that the model is not limited to memorizing the species included in the post-training data. Instead, the Auditor–Synthesizer-based CoT appears to teach a general reasoning procedure: retrieve the diagnostic traits of the prompted species, compare those traits with visible plant evidence, account for missing or occluded traits, and return the corresponding bounding boxes. This behavior is important for agricultural applications because growers may need to query different weed species depending on crop, region, or field history.

Although these target species were absent from the SFT and GRPO post-training data, WeedExpert-R1 was still to localize the prompted target instances in the selected examples This provides qualitative evidence of zero-shot grounding with respect to the post-training data and suggests that the open-vocabulary formulation can support species-level grounding beyond the fixed set of species used for SFT and GRPO. The results also indicate that WeedExpert-R1 can combine botanical knowledge from the base MLLM with the structured reasoning procedure learned through Auditor–Synthesizer CoT supervision. For common or well-described plants, such as volunteer corn and sugar beet, this ability likely benefits from botanical knowledge already encoded in the base MLLM during pretraining. For unseen species such as Canada thistle, WeedExpert-R1 does not rely on species-specific post-training examples. Instead, it applies a reusable reasoning procedure learned from botanical reasoning CoTs: forming a trait-based search profile from the species prompt, comparing visible plant evidence with that profile, accounting for missing or occluded traits, and linking the evidence to the correct image regions. This ability to reason about missing traits is especially important for top-down field imagery, where stems, petioles, or leaf attachments may be hidden by canopy structure, crop leaves, or camera viewpoint. The final example further illustrates the flexibility of prompt-based grounding. The same image was tested with different prompts, asking the model to ground either \textit{Tribulus terrestris} or \textit{Beta vulgaris}. WeedExpert-R1 adjusted its reasoning according to the prompted species, applied the corresponding botanical trait profile, and returned the target instances associated with each prompt. This demonstrates a practical use case in which a grower or agronomist can query the model for a species of interest without redefining a fixed detector class list.

\textbf{Practical Deployment Implications.} From a practical deployment perspective, our WeedExpert-R1 model can be used with field images collected by ground-based cameras, robotic platforms, or individual frames extracted from field videos. Because the model operates on RGB image inputs and receives the target species as a natural-language prompt, it is compatible with common field-imaging systems used in precision agriculture. This capability is particularly relevant for robotic weed scouting and site-specific weed management, where images are collected continuously under variable field conditions and the target species may differ by crop, region, or season.

\section{Conclusion}
This work presents \textbf{WeedExpert-R1}, a multimodal reasoning model that incentivizes visually grounded botanical reasoning for species-level weed identification and grounding through verifiable rewards under the R1-style training paradigm. The central contribution is an Auditor–Synthesizer CoT synthesis framework that converts expert botanical trait knowledge into structured, visually grounded reasoning traces for MLLM post-training. Specifically, the proposed pipeline couples a human-curated botanical trait dictionary, covering leaf shape, margin, petiole, and stem morphology, with an Auditor–Synthesizer LLM workflow. The Auditor verifies visible traits for each weed instance, and the Synthesizer aggregates these observations into image-level botanical CoTs. This design provides quality-controlled reasoning data for cold-start SFT and helps establish a structured botanical reasoning pattern in the base MLLM.

Building on this cold-start model, GRPO with four verifiable rewards, including format compliance, bounding-box IoU, count consistency, and reasoning-length control, is applied to further refine grounding precision. Experimental results demonstrate that WeedExpert-R1-4B outperforms both frontier proprietary models (GPT-5.4, Gemini-3.1-Pro) and larger open-source baselines (Qwen3-VL-30B-Instruct, Gemma-4-31B-it), achieving an average improvement of approximately 35 percentage points across five evaluation metrics on 37 weed species. Qualitative examples on selected species absent from the SFT and GRPO post-training data further suggest the potential of WeedExpert-R1 for open-vocabulary weed grounding. Overall, these results indicate that domain-specific botanical reasoning supervision and verifiable reward-based reinforcement learning can substantially improve MLLM-based weed grounding, and they provide a promising foundation for future deployment in precision agriculture systems, including robotic and edge-based weed management.

Our future work will focus on improving CoT quality and enabling field deployment. Higher-quality botanical CoT generation will be explored by incorporating multi-instance comparison, which can help distinguish visually similar weed species within the same image, and visual tool use, such as zoom-in or region-level inspection, to support finer trait verification.In the near term, our WeedExpert-R1 model could also be deployed through a server-side inference system, where growers, crop scouts, or agronomists upload field images through a web or mobile interface and query species of interest. The model would run on GPU-backed computing resources and return localized target plants together with botanical reasoning traces. In the longer term, such reasoning traces can equip the base MLLM with diverse agentic capabilities, providing a foundation for subsequent agentic reinforcement learning. Because such agentic capabilities may require larger MLLMs, such as 30B-scale models, our future work will also investigate on-policy distillation to transfer the reasoning and grounding ability of larger models into compact student models. This step is important for deploying WeedExpert-R1-like models on resource-constrained robotic platforms and intelligent edge devices for in-field weed scouting and precision weed management.

\section{Acknowledgments}
The authors gratefully acknowledge the support of the Beet Sugar Development Foundation (Award No. 820), U.S. National Science Foundation Experiential Learning for Emerging and Novel Technologies (ExLENT) Program (Award No. 2322535), NSF-EPSCoR (Award No. 2420405), USDA-CBG (Award No. 2024-38821-42099), and USDA-NIFA (Award No. 2024-67021-41534).





%
%





\printcredits

\bibliographystyle{cas-model2-names}

\bibliography{cas-refs}

@article{soltani2016potential,
  title={Potential corn yield losses from weeds in North America},
  author={Soltani, Nader and Dille, J Anita and Burke, Ian C and Everman, Wesley J and VanGessel, Mark J and Davis, Vince M and Sikkema, Peter H},
  journal={Weed Technology},
  volume={30},
  number={4},
  pages={979--984},
  year={2016},
  publisher={Cambridge University Press}
}

@article{grattafiori2024llama,
  title={The llama 3 herd of models},
  author={Grattafiori, Aaron and Dubey, Abhimanyu and Jauhri, Abhinav and Pandey, Abhinav and Kadian, Abhishek and Al-Dahle, Ahmad and Letman, Aiesha and Mathur, Akhil and Schelten, Alan and Vaughan, Alex and others},
  journal={arXiv preprint},
  volume={arXiv:2407.21783},
  year={2024}
}

@article{gauba2026agmmu,
  title={AgMMU: a comprehensive agricultural multimodal understanding benchmark},
  author={Gauba, Aruna and Pi, Irene and Man, Yunze and Pang, Ziqi and Adve, Vikram and Wang, Yu-Xiong},
  journal={Advances in Neural Information Processing Systems},
  volume={38},
  year={2026}
}

@inproceedings{awais2025agrogpt,
  title={Agrogpt: Efficient agricultural vision-language model with expert tuning},
  author={Awais, Muhammad and Alharthi, Ali Husain Salem Abdulla and Kumar, Amandeep and Cholakkal, Hisham and Anwer, Rao Muhammad},
  booktitle={2025 IEEE/CVF Winter Conference on Applications of Computer Vision (WACV)},
  pages={5687--5696},
  year={2025},
  organization={IEEE}
}

@article{liu2023visual,
  title={Visual instruction tuning},
  author={Liu, Haotian and Li, Chunyuan and Wu, Qingyang and Lee, Yong Jae},
  journal={Advances in neural information processing systems},
  volume={36},
  pages={34892--34916},
  year={2023}
}

@book{lawrence2021herbicide,
  title={Herbicide Options for Control of Glyphosate-Resistant Weeds in Sugar Beet},
  author={Lawrence, Nevin and Kniss, Andrew},
  year={2021},
  publisher={University of Nebraska-Lincoln, Extension Division}
}

@article{jaech2024openai,
  title={Openai o1 system card},
  author={Jaech, Aaron and Kalai, Adam and Lerer, Adam and Richardson, Adam and El-Kishky, Ahmed and Low, Aiden and Helyar, Alec and Madry, Aleksander and Beutel, Alex and Carney, Alex and others},
  journal={arXiv preprint},
  volume={arXiv:2412.16720},
  year={2024}
}

@article{zhang2026agri,
  title={Agri-R1: Agricultural Reasoning for Disease Diagnosis via Automated-Synthesis and Reinforcement Learning},
  author={Zhang, Wentao and Xu, Mingkun and Zhang, Qi and Li, Shangyang and Wong, Derek F and Wang, Lifei and Yang, Yangchao and Lu, Lina and Fang, Tao},
  journal={arXiv preprint},
  volume={arXiv:2601.04672},
  year={2026}
}

@inproceedings{liu2024multimodal,
  title={A multimodal benchmark dataset and model for crop disease diagnosis},
  author={Liu, Xiang and Liu, Zhaoxiang and Hu, Huan and Chen, Zezhou and Wang, Kohou and Wang, Kai and Lian, Shiguo},
  booktitle={European Conference on Computer Vision},
  pages={157--170},
  year={2024},
  organization={Springer}
}

@article{hong2025deepeyesv2,
  title={Deepeyesv2: Toward agentic multimodal model},
  author={Hong, Jack and Zhao, Chenxiao and Zhu, ChengLin and Lu, Weiheng and Xu, Guohai and Yu, Xing},
  journal={arXiv preprint},
  volume={arXiv:2511.05271},
  year={2025}
}

@article{miranda2024crop,
  title={Crop safety and control of acetolactate synthase inhibitor-resistant Palmer amaranth (Amaranthus palmeri) with very long-chain fatty acid-inhibiting herbicides in dry edible bean},
  author={Miranda, Joshua WA and Jhala, Amit J and Bradshaw, Jeffrey and Lawrence, Nevin C},
  journal={Frontiers in Agronomy},
  volume={6},
  pages={1401865},
  year={2024},
  publisher={Frontiers Media SA}
}

@article{akuoko2026evaluating,
  title={Evaluating weed control in dicamba, glyphosate, and glufosinate-resistant sugar beet in the western United States},
  author={Akuoko, Abraham and Adjesiwor, Albert and Felix, Joel and Kniss, Andrew and Lawrence, Nevin},
  journal={Weed Technology},
  volume={40},
  pages={e7},
  year={2026},
  publisher={Cambridge University Press}
}

@inproceedings{zhao2024detrs,
  title={Detrs beat yolos on real-time object detection},
  author={Zhao, Yian and Lv, Wenyu and Xu, Shangliang and Wei, Jinman and Wang, Guanzhong and Dang, Qingqing and Liu, Yi and Chen, Jie},
  booktitle={Proceedings of the IEEE/CVF conference on computer vision and pattern recognition},
  pages={16965--16974},
  year={2024}
}

@article{zhang2022dino,
  title={Dino: Detr with improved denoising anchor boxes for end-to-end object detection},
  author={Zhang, Hao and Li, Feng and Liu, Shilong and Zhang, Lei and Su, Hang and Zhu, Jun and Ni, Lionel M and Shum, Heung-Yeung},
  journal={arXiv preprint},
  volume={arXiv:2203.03605},
  year={2022}
}

@article{khanam2024yolov11,
  title={Yolov11: An overview of the key architectural enhancements},
  author={Khanam, Rahima and Hussain, Muhammad},
  journal={arXiv preprint},
  volume={arXiv:2410.17725},
  year={2024}
}

@article{yang2026weedcam,
  title={WeedCAM: An edge-computing camera system for multi-species weed detection in sugar beet production fields},
  author={Yang, Zonglin and Liang, Wei-Zhen and Lawrence, Nevin and Qiao, Xin and Riggan, Benjamin and Harveson, Robert and Chiang, Chi-En and Oboamah, Joseph and Andjawo, Diwenitissiou Philipine},
  journal={Computers and Electronics in Agriculture},
  volume={244},
  pages={111498},
  year={2026},
  publisher={Elsevier}
}

@article{wang2026resource,
  title={Resource-Constrained UAV-Based Weed Detection for Site-Specific Management on Edge Devices},
  author={Wang, Linyuan and Yao, Haibo and Tseng, Te-Ming and Betitame, Kelvin and Sun, Xin and Huang, Hanbo and Chen, Dong},
  journal={arXiv preprint},
  volume={arXiv:2604.23442},
  year={2026}
}

@inproceedings{liu2025visual,
  title={Visual-rft: Visual reinforcement fine-tuning},
  author={Liu, Ziyu and Sun, Zeyi and Zang, Yuhang and Dong, Xiaoyi and Cao, Yuhang and Duan, Haodong and Lin, Dahua and Wang, Jiaqi},
  booktitle={Proceedings of the IEEE/CVF International Conference on Computer Vision},
  pages={2034--2044},
  year={2025}
}

@article{zheng2025deepeyes,
  title={Deepeyes: Incentivizing ``Thinking with Images'' via Reinforcement Learning},
  author={Zheng, Ziwei and Yang, Michael and Hong, Jack and Zhao, Chenxiao and Xu, Guohai and Yang, Le and Shen, Chao and Yu, Xing},
  journal={arXiv preprint},
  volume={arXiv:2505.14362},
  year={2025}
}

@article{liu2025visionreasoner,
  title={Visionreasoner: Unified visual perception and reasoning via reinforcement learning},
  author={Liu, Yuqi and Qu, Tianyuan and Zhong, Zhisheng and Peng, Bohao and Liu, Shu and Yu, Bei and Jia, Jiaya},
  journal={arXiv e-prints},
  pages={arXiv--2505},
  year={2025}
}

@inproceedings{liu2024grounding,
  title={Grounding dino: Marrying dino with grounded pre-training for open-set object detection},
  author={Liu, Shilong and Zeng, Zhaoyang and Ren, Tianhe and Li, Feng and Zhang, Hao and Yang, Jie and Jiang, Qing and Li, Chunyuan and Yang, Jianwei and Su, Hang and others},
  booktitle={European conference on computer vision},
  pages={38--55},
  year={2024},
  organization={Springer}
}

@article{schulman2017proximal,
  title={Proximal policy optimization algorithms},
  author={Schulman, John and Wolski, Filip and Dhariwal, Prafulla and Radford, Alec and Klimov, Oleg},
  journal={arXiv preprint},
  volume={arXiv:1707.06347},
  year={2017}
}

@article{schulman2015high,
  title={High-dimensional continuous control using generalized advantage estimation},
  author={Schulman, John and Moritz, Philipp and Levine, Sergey and Jordan, Michael and Abbeel, Pieter},
  journal={arXiv preprint},
  volume={arXiv:1506.02438},
  year={2015}
}

@article{shao2024deepseekmath,
  title={Deepseekmath: Pushing the limits of mathematical reasoning in open language models, 2024},
  author={Shao, Zhihong and Wang, Peiyi and Zhu, Qihao and Xu, Runxin and Song, Junxiao and Bi, Xiao and Zhang, Haowei and Zhang, Mingchuan and Li, YK and Wu, Yang and others},
  journal={URL https://arxiv. org/abs/2402.03300},
  volume={2},
  number={3},
  pages={5},
  year={2024}
}

@article{guo2025deepseek,
  title={Deepseek-r1: Incentivizing reasoning capability in llms via reinforcement learning},
  author={Guo, Daya and Yang, Dejian and Zhang, Haowei and Song, Junxiao and Wang, Peiyi and Zhu, Qihao and Xu, Runxin and Zhang, Ruoyu and Ma, Shirong and Bi, Xiao and others},
  journal={arXiv preprint},
  volume={arXiv:2501.12948},
  year={2025}
}

@misc{openai2026gpt54,
  title={{GPT-5.4 Thinking System Card}},
  author={{OpenAI}},
  year={2026},
  month={March},
  howpublished={\url{https://openai.com/index/gpt-5-4-thinking-system-card/}},
  note={Accessed: 2026-04-09}
}

@techreport{deepmind2026gemini31,
  title={{Gemini 3.1 Pro System Card Updates}},
  author={{DeepMind}},
  year={2026},
  month={February},
  institution={{Google}},
  howpublished={\url{https://deepmind.google/research/}},
  note={Accessed: 2026-04-09}
}

@misc{anthropic2026opus46,
  title={{Claude Opus 4.6 Overview and System Capabilities}},
  author={{Anthropic}},
  year={2026},
  month={February},
  howpublished={\url{https://www.anthropic.com/claude/opus}},
  note={Accessed: 2026-04-09}
}

@article{kuhn1955hungarian,
  title={The Hungarian method for the assignment problem},
  author={Kuhn, Harold W},
  journal={Naval research logistics quarterly},
  volume={2},
  number={1-2},
  pages={83--97},
  year={1955},
  publisher={Wiley Online Library}
}

@article{deng2025weed,
  title={Weed image augmentation by ControlNet-added stable diffusion for multi-class weed detection},
  author={Deng, Boyang and Lu, Yuzhen},
  journal={Computers and Electronics in Agriculture},
  volume={232},
  pages={110123},
  year={2025},
  publisher={Elsevier}
}

@article{dang2023yoloweeds,
  title={YOLOWeeds: A novel benchmark of YOLO object detectors for multi-class weed detection in cotton production systems},
  author={Dang, Fengying and Chen, Dong and Lu, Yuzhen and Li, Zhaojian},
  journal={Computers and Electronics in Agriculture},
  volume={205},
  pages={107655},
  year={2023},
  publisher={Elsevier}
}

@inproceedings{rahman2022deep,
  title={Deep neural networks for weed detections towards precision weeding},
  author={Rahman, Abdur and Lu, Yuzhen and Wang, Haifeng},
  booktitle={2022 ASABE Annual International Meeting},
  pages={1},
  year={2022},
  organization={American Society of Agricultural and Biological Engineers}
}

@article{upadhyay2025weed,
  title={Weed-crop dataset in precision agriculture: Resource for AI-based robotic weed control systems},
  author={Upadhyay, Arjun and Mahecha, Maria Villamil and Mettler, Joseph and Howatt, Kirk and Aderholdt, William and Ostlie, Michael and Sun, Xin and others},
  journal={Data in Brief},
  volume={60},
  pages={111486},
  year={2025},
  publisher={Elsevier}
}

@article{wang2022weed25,
  title={Weed25: A deep learning dataset for weed identification},
  author={Wang, Pei and Tang, Yin and Luo, Fan and Wang, Lihong and Li, Chengsong and Niu, Qi and Li, Hui},
  journal={Frontiers in Plant Science},
  volume={13},
  pages={1053329},
  year={2022},
  publisher={Frontiers Media SA}
}

@article{bai2025qwen3,
  title={Qwen3-vl technical report},
  author={Bai, Shuai and Cai, Yuxuan and Chen, Ruizhe and Chen, Keqin and Chen, Xionghui and Cheng, Zesen and Deng, Lianghao and Ding, Wei and Gao, Chang and Ge, Chunjiang and others},
  journal={arXiv preprint},
  volume={arXiv:2511.21631},
  year={2025}
}

@misc{google2026gemma4,                                                                                                                      
    title = {Gemma 4 Model Card},                                                                                                              
    author = {{DeepMind}},                                                                                                              
    year = {2026},                                                                                                                             
    howpublished = {\url{https://ai.google.dev/gemma/docs/core/model_card_4}},                                                                 
    note = {Accessed: 2026-05-01}                                                                                                              
  }

@article{he2023grec,
  title={Grec: Generalized referring expression comprehension},
  author={He, Shuting and Ding, Henghui and Liu, Chang and Jiang, Xudong},
  journal={arXiv preprint},
  volume={arXiv:2308.16182},
  year={2023}
}

@misc{roboflow,
  author = {{Roboflow}},
  title = {Roboflow},
  howpublished = {\url{https://roboflow.com}},
  year = {2026},
  note = {Accessed: 2026-05-01}
}

@article{shen2025vlm,
  title={Vlm-r1: A stable and generalizable r1-style large vision-language model},
  author={Shen, Haozhan and Liu, Peng and Li, Jingcheng and Fang, Chunxin and Ma, Yibo and Liao, Jiajia and Shen, Qiaoli and Zhang, Zilun and Zhao, Kangjia and Zhang, Qianqian and others},
  journal={arXiv preprint},
  volume={arXiv:2504.07615},
  year={2025}
}

@article{fu2025reward,
  title={Reward shaping to mitigate reward hacking in rlhf},
  author={Fu, Jiayi and Zhao, Xuandong and Yao, Chengyuan and Wang, Heng and Han, Qi and Xiao, Yanghua},
  journal={arXiv preprint},
  volume={arXiv:2502.18770},
  year={2025}
}



\end{document}